\documentclass{article}

\usepackage{microtype}
\usepackage{graphicx}
\usepackage{subfigure}
\usepackage{booktabs} % for professional tables
\usepackage{algorithmic}
\usepackage{hyperref}
\usepackage{tabularx}
\usepackage{multirow}

\newcommand{\norm}[1]{\left\lVert#1\right\rVert} 
\renewcommand{\cite}{\citep}

\usepackage[accepted]{icml2025}

\usepackage{amsmath}
\usepackage{amssymb}
\usepackage{mathtools}
\usepackage{amsthm}
\usepackage{enumitem}
\usepackage{multirow}
\usepackage{mathtools}
\usepackage[utf8]{inputenc}
\usepackage{adjustbox}
\usepackage{graphicx}
\usepackage{tikz}
\usepackage{xcolor}
\usepackage[capitalize,noabbrev]{cleveref}

\theoremstyle{plain}

\theoremstyle{definition}

\theoremstyle{remark}

\usepackage[textsize=tiny]{todonotes}

\icmltitlerunning{Instance-Level Difficulty: A Missing Perspective in Machine Unlearning}

\begin{document}

\twocolumn[
\icmltitle{Position: Instance-Level Difficulty: A Missing Perspective \\ in Machine Unlearning}



\icmlsetsymbol{equal}{*}

\begin{icmlauthorlist}
\icmlauthor{Hammad Rizwan}{equal,yyy}
\icmlauthor{Mahtab Sarvmaili}{equal,yyy}
\icmlauthor{Hassan Sajjad}{yyy}
\icmlauthor{Ga Wu}{yyy}
\end{icmlauthorlist}

\icmlaffiliation{yyy}{Department of Computer Science, Dalhousie University, Canada}

\icmlcorrespondingauthor{Hammad Rizwan}{hammad.rizwan@dal.ca}
\icmlcorrespondingauthor{Mahtab Sarvmaili}{mahtab.sarvmaili@dal.ca}

\icmlkeywords{Machine Learning, ICML}

\vskip 0.3in
]



\printAffiliationsAndNotice{\icmlEqualContribution} % otherwise use the standard text.

\begin{abstract}
% In response to recent privacy protection regulations, machine unlearning has attracted great interest in the research community. However, existing studies often demonstrate their approaches' effectiveness by measuring the overall unlearning success rate rather than evaluating the difficulty of unlearning specific training samples, leaving the universal feasibility of the unlearning operation unexplored. This paper proposes a novel method to quantify the difficulty of machine unlearning for a single sample by taking model and data distribution into account jointly. Specifically, we propose several heuristics to understand the condition of a successful unlearning operation on data points, explore difference in unlearning difficulty over training data points, and suggest a potential ranking mechanism for identifying the most challenging samples to unlearn.  In particular, we note Kernelized Stein Discrepancy (KSD), a parameterized kernel function tailored to each model and dataset, is an effective heuristic to tell the difficulty of unlearning a data sample. We demonstrate our discovery by including multiple classification tasks and existing machine unlearning algorithms, highlighting the practical feasibility of unlearning operations across different scenarios.

%In light of recent privacy regulations, machine unlearning has attracted significant attention in the research community. 

%However, 
%Current deep machine unlearning studies predominantly assess the overall success of unlearning approaches
Current research on deep machine unlearning primarily focuses on improving or evaluating the overall effectiveness of unlearning methods while overlooking the varying difficulty of unlearning individual training samples. As a result, the broader feasibility of machine unlearning remains under-explored. This paper studies the cruxes that make machine unlearning difficult through a thorough instance-level unlearning performance analysis over various unlearning algorithms and datasets. In particular, we summarize four factors that make unlearning a data point difficult, and we empirically show that these factors are independent of a specific unlearning algorithm but only relevant to the target model and its training data. Given these findings, we argue that machine unlearning research should pay attention to the instance-level difficulty of unlearning.  
\end{abstract}

\section{Introduction}

% Machine unlearning (MU) was first introduced by \cite{cao2015towards} to address the challenge of enabling machine learning (ML) systems to completely forget specific pieces of training data and revert their effects on the models while preserving the models' performance on the remaining data. 
Machine Unlearning~(MU)~\cite{cao2015towards} refers to a process that enables machine learning (ML) models to remove specific training data and revert corresponding data influence on the trained models while preserving the models' generalization. As many countries and territories have promulgated their Right to be Forgotten regulations~\footnote{CCPA in California, GDPR in Europe, PIPEDA in Canada,
LGPD in Brazil, and NDBS in Australia.}, entitling individuals to revoke their authorization to use their data for machine learning (ML) model training, the demand of MU raised significant interest in the ML research community, leading to various types of unlearning approaches, often achieved by either data reorganization \cite{graves2021amnesiac, gupta2021adaptive, tarun2023fast} or model manipulation \cite{pmlr-v119-guo20c, warnecke2021machine}.

Although existing machine unlearning studies vary based on diverse theoretical foundations, they often rely on similar quantitative performance evaluation metrics, including 1) Data Erasure Completeness, 2) Unlearning Time Efficiency, 3) Resource Consumption, and 4) Privacy Preservation~\cite{xu2024machine, yang2023survey, shaik2023exploring}. It is an implicit consensus that the variants of the above metrics suffice for comparing the performance of MU methods from various perspectives.
However, we highlight that they often fall short in assessing the effectiveness of data removal requests for individual data points, resulting in a discrepancy between actual unlearning outcomes and performance expectations in real-world applications. Indeed, the difficulty of unlearning individual data points exhibits significant variability~\cite{marchant2022hard,pawelczyk2024machine, zhao2024makes} that should not be overlooked; Some data points are inherently harder to unlearn than others, where such variability may stem from intrinsic factors, such as the augmented data distribution under a trained machine learning model, regardless of the specific MU algorithm applied. All above leaves an inescapable question: how to perceive and quantify instance-level unlearning difficulty?

This paper investigates the difficulty (or even feasibility) of machine unlearning by logging and analyzing the outcome of unlearning operations on each training data point from a trained model empirically. Through thorough analysis of four MU algorithms from different algorithm families and three benchmark datasets, we identify four factors that pose the challenge to unlearning operations, regardless of the choice of specific MU algorithms. Further analysis reveals that the four identified factors capture different types or definitions of unlearning difficulty, as the challenging samples identified by each factor exhibit notable diversity. {\bf Given the above findings, we argue that
machine unlearning research should pay attention
to the instance-level difficulty (or even feasibility) of unlearning.}

The four identified factors above can quantify unlearning difficulty but are impractical for predicting outcomes before execution. This gap highlights the need for a unified difficulty index to forecast unlearning results and reduce computational costs (research gap). Currently, no proven effective index exists in the literature.

\section{Preliminaries}
\subsection{Objective of Machine Unlearning (Definition)}
\label{sec:mu_definition}
Machine Unlearning (MU) is the process of removing specific subsets of training data, along with their influence, from a trained model~\cite{cao2015towards, bourtoule2021machine}. Ideally, the unlearned model should perform identically to a model trained from scratch on a pruned dataset, where the data targeted for removal has been excluded. However, in practice, quantifying the performance of unlearning operations based on the above criterion is challenging, as it requires a retrained model to serve as a reference. Alternatively, existing MU research often measures the success of unlearning operation through two surrogate indices, namely 1) Model Utility Retention and 2) Unlearning Effectiveness. 

Consider a training dataset $D_t = \{(\mathbf{x}_{i}, y_{i})\}$ ($t$ is refering to training data) comprising $n$ samples, where $\mathbf{x}_{i}$ and $y_{i}$ represent the $i^{th}$ data's features and corresponding label respectively. We define two subsets of the dataset for clarity as follows: Let $D_f \subseteq D_t$ denote the subset of data designated to be forgotten (a.k.a {\it forget set}), and $D_r \subseteq D_t$ denote the remaining data (a.k.a {\it remaining set}), such that $D_f \cup D_r = D_t$ and $D_f \cap D_r = \emptyset$.

Given an target predictive model $f_{\theta}$ with parameters $\theta$, the common expectation of machine unlearning operation are of adjusting $\theta$ to a modified parameter set $\vartheta$ such that:
\begin{enumerate}[leftmargin=4mm, itemsep=-1mm, topsep=-1mm]
    \item Increasing of model's error on the forget set $\mathcal{L}_{\vartheta}(D_f)$.
    \item Maintaining original model's error on the remainingset $D_{r}$  such that $
        \norm{\mathcal{L}_{\theta}(D_r) - \mathcal{L}_{\vartheta}(D_r)} < \epsilon,
    %\label{eq:preservation_1}
    $
    where $\epsilon$ denotes a tolerable performance degradation threshold and $\mathcal{L}$ denotes the loss.

\end{enumerate}

\subsection{Research Track of Machine Unlearning}
\label{sec:mu_algorithms}
The simplest solution for unlearning is to retrain the model from scratch using the remaining data after removing the forget data, but this process is resource-intensive even with partial retraining techniques~\cite{bourtoule2021machine}. To reduce computational costs, approximations like Fine-Tuning~\cite{warnecke2021machine, golatkar2020eternal} continue training on the remaining data ($D_r$) to naturally diminish the influence of forget data ($D_f$). However, fine-tuning-based approaches can significantly alter model parameters and become inefficient as $D_r$ grows. In contrast, Gradient Ascent (GA)\cite{graves2021amnesiac} adjusts weights to increase the model's error on $D_f$, though this often impacts predictive performance. NegGrad+\cite{kurmanji2024towards} (in this paper, from this point forward, we will refer to NegGrad+ simply as NegGrad) addresses the weakness above by combining fine-tuning on $D_r$ with GA on $D_f$ for balanced unlearning. Aside of the above gradient based objectives, we also see much more advanced objectives, such as SCRUB, Infuence Unlearning, and SalUn. In particular, SCRUB uses a student-teacher optimization where the student aligns with the teacher on $D_r$ but diverges on $D_f$. Influence Unlearning\cite{izzo2021approximate} employs Influence Functions (IF) with WoodFisher Hessian approximation to estimate parameter changes caused by removing data. Saliency Unlearning (SalUn)~\cite{fan2023salun} uses relabeling techniques, fine-tuning on a relabeled dataset ($D_{\text{relabel}}$) and optimizing only salient parameters identified by gradient updates to shift class predictions effectively.

Recent studies have frequently utilized the Newton update as a fundamental step for removing data influence \cite{pmlr-v119-guo20c, golatkar2020eternal, peste2021ssse, sekhari2021remember}. These methods typically leverage the Fisher Information Matrix (FIM) to gauge the sensitivity of the model's output to perturbations in its parameters. %For example, Fisher Forgetting \cite{golatkar2020eternal} employs a scrubbing approach where noise is added to parameters based on their relative importance in distinguishing the forget set from the remaining data set. Mehta et al. \cite{mehta2022deep} employs conditional independence coefficient to identify sufficient sets of parameters for targeted unlearning. 

\subsection{Quantifying Performance of Unlearning}
%{\color{red} Wuga: please put citations here. We need to fill up the second page with those.}

The majority of the literature on machine unlearning primarily concentrates on the development of unlearning algorithms or unlearning approximation techniques for selectively forgetting data from a trained model. As such, the corresponding evaluation metrics are designed to favor the performance difference between algorithms on a highly aggregated level (e.g. success rate). An implicit consensus underlying much of this research is that unlearning operations are universally feasible for all data points within a dataset, where effectiveness of unlearning will behave consistently across different datasets. 
%This assumption often overlooks the potential variability in unlearning efficacy due to differences in data characteristics or model dependencies, suggesting a need for more nuanced studies that evaluate the specific conditions under which MU can be effectively implemented. %\textcolor{red}{Recently, one line work \cite{thudi2022necessity} has questioned the performance of unlearning approximation algorithms w.r.t the exact unlearning and whether or not these methods can successfully prove the
% absence of certain data points during training. From this recent research, we can observe that absence of a comprehensive study that comprehensively investigates the feasibility of unlearning was present.}
In fact, by scanning MU research literature, we note there were research~\cite{thudi2022necessity} that questioned whether unlearning approximations can reliably emulate exact unlearning and prove the absence of specific data points during training, highlighting the lack of comprehensive studies on the feasibility of unlearning. Similarly, \citep{fan2025challenging} explored challenges like worst-case forget sets through adversarial unlearning and the interplay between $D_r$, $D_f$, and model memorization strength. Unfortunately, none of these studies touch the base of instance-level unlearning difficulties and fail to perceive the factors attributing to the complexity of using unlearning algorithms in practice, signalling the need for more nuanced investigations.

% \textcolor{red}{MAHTAB: PLEASE CHECK THE FOLLOWING PART. WE SHOULD JUSTIFY WHY WE ARE FOCUSING ON SEVERAL ALGORITHMS.}
% \textcolor{blue}{The MU community is increasingly focused on understanding the challenges and limitations associated with unlearning algorithms. Recent studies have explored obtaining worst-case forget sets through the lens of adversarial unlearning \cite{fan2025challenging}, as well as the entanglement of $D_r$ and $D_f$ and the model's memorization strength of the data. However, both approaches overlook the challenges associtated with the unlearning algorithms which can be significanlty exacerbated for individual unlearning, and most importantly inestigation of other contributing factors for unlearning difficulty and most importantly.}

% \begin{figure*}[t]
%     \centering
%         \includegraphics[width=\textwidth]{icml2025/figs/single_unlearn_figures/drop_perdictive_prob_forget_mnist.pdf}
%         \caption{Testing image}
%         \label{fig:common_neurons_box_plot}
%         \vspace{-0.4cm}
% \end{figure*}

    \begin{figure*}[t]
    \centering
        \includegraphics[width=\textwidth]{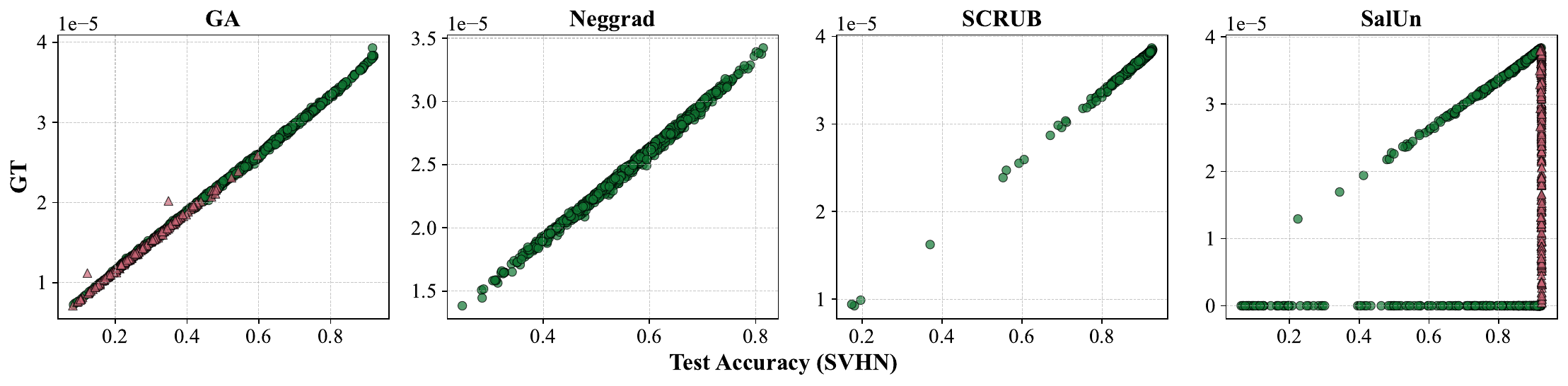}
        \vspace{-5mm}
        \caption{Effectiveness of using Tolerance of Preference Shift (TPS) as an index of unlearning difficulty. 
        %\textcolor{red}{(Top) Accuracy of unlearned model on test data. (Bottom) Drop of model's loss on test set post-unlearning.}. 
        For the four different unlearning algorithms, we note there is a consistent positive alignment between TPS and empirical unlearning outcome (GT). The experiments are conducted on ResNet-18 model trained on SVHN dataset. }
        \vspace{-5mm}
        \label{fig:acc_loss_test_svhn}
    \end{figure*}
    
\section{Factors that Affect Difficulty of Unlearning}
\label{sec:unlearning_factors}
In this section, we summarize the possible factors that impact the effectiveness of machine unlearning with corresponding data analysis and intuition justifications. The factors discussed here are those commonly used in existing studies in the machine unlearning literature, where they are often implicitly assumed without explicit justification. 

\subsection{Ground-truth Unlearning Outcome}
To facilitate our analysis, we first quantify the success of unlearning operation such that reflects the common consensus described in Section~\ref{sec:mu_definition}. 

Considering unlearning an individual data point $(\mathbf{x}_f,y_f)\in D_f$ (where $|D_f| = 1$), we define the success of the unlearning operation (ground-truth or GT) as a Harmonic average (FP-Score) over model's performance shift on the forget set and remaining (or test) set in the form
\begin{equation}
    \text{GT}(\mathcal{A},D_f) = \frac{\lambda * U * F}{\lambda*U+F},
\end{equation}
where $U$ denotes the difference between the predicted probability of test set
\vspace{-3mm}
\begin{equation}
    U= 1 - \frac{1}{|D_s|} \sum_i^{|D_s|}\bigg[{f_\vartheta} (\mathbf{x}_i)_{y_i} - {f_\theta}(\mathbf{x}_i)_{y_i}\bigg],
\end{equation}
$F$ denotes the difference between the predicted
probability of forget set 
\begin{equation}
    F =  {f_\vartheta}(\mathbf{x}_f)_{y_f} - {f_\theta}(\mathbf{x}_f)_{y_f},
\end{equation}
and $\lambda$ denotes the a balancing factor such that $\lambda = |F|/|U|$. We use $\mathcal{A}$ to denote the unlearning algorithm used for computing the GT.

The GT score intuitively captures the two unlearning objectives outlined in Section~\ref{sec:mu_definition} using a single metric. Since the harmonic average is dominated by smaller values, it is particularly sensitive to the worst-case outcomes in machine unlearning. This ensures that both poor generalization to test sets and failures in forgetting are treated as indicators of unsuccessful unlearning. More concretely, a larger $ \text{GT} $ implies $ U \to 1 $ and $ F \to 1 $, indicating the unlearning of easy samples. Conversely, a smaller $ \text{GT} $ suggests the unlearning of more difficult samples. It is worth noting that using the Harmonic average score as performance measurement is a common practice in the modern AI\&ML literature~\cite{song2024measuring}.
    
\subsection{Scope of Empirical Analysis}
Within the families of unlearning algorithms summarized in Section~\ref{sec:mu_algorithms}, we focus our analysis on GA, NegGrad, SCRUB, and SalUn due to their broad applicability. Prior studies~\cite{liu2024model, ding2024fine} indicate that Influence Unlearning and Fine-Tuning (FT) are ineffective for unlearning individual samples. Although FT aims to preserve the model’s utility on the remaining data, it struggles to forget targeted data points, a challenge that becomes even more pronounced when a large number of similar samples remain in the retained dataset~\cite{ding2024fine}. Similarly, Influence Unlearning is impractical for individual sample removal, as it requires computing the Hessian-vector product twice per sample, making it infeasible for real-world applications.

In the main paper, we present our analysis results using plots from models trained on the SVHN dataset due to space constraints. However, our study encompasses multiple datasets and experimental settings, and the observed trends remain consistent across these variations. Please refer to the Appendix for more results.

\begin{figure*}[t]
\centering
    \includegraphics[width=\textwidth]{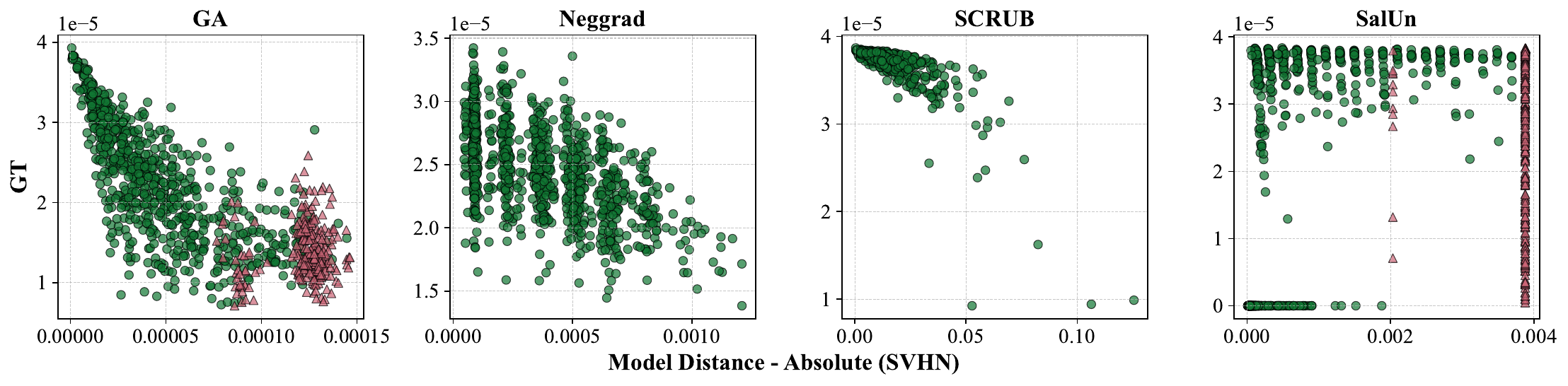}
   \vspace{-5mm}
        \caption{Effectiveness of using Distance of Preference Shift (DPS) as an index of unlearning difficulty. 
        %\textcolor{red}{(Top) Absolute average layer-wise distance of the model's weights pre- and post-unlearning. (Bottom) KL-divergence of model's parameters before and after unlearning.}
        For the tree out of four different unlearning algorithms, we observe a negative alignment between DPS and empirical unlearning outcome (GT). For SalUn, there is no clear correlation between DPS and GT. The experiments are conducted on ResNet-18 model trained on SVHN dataset. }
    \label{fig:model_parm_shift_abs}
    \vspace{-3mm}
\end{figure*}
    
\subsection{Analyzing Difficulty of Unlearning}
\label{sec:six_factors}
We now present our analytical results, which examine the alignment between the empirical outcomes of the unlearning operation (GT) and the factors commonly considered as influencing the difficulty of unlearning. Specifically, we evaluate the reliability of these factors in indicating the challenges associated with unlearning.

\subsubsection{Tolerance of Performance Shift}

The Tolerance of Performance Shift (TPS) quantifies the difficulty of unlearning by imposing a strict requirement for guaranteed unlearning~\cite{liu2024survey}. Specifically, an unlearning algorithm must ensure a decision flip for the forget set, often requiring a substantially large unlearning step size. Formally, the TPS denotes the solution of the following optimization task
\begin{equation}
\begin{aligned}
    &\min \frac{1}{|D_s|} \sum_i^{|D_s|}\bigg[{f_\vartheta} (\mathbf{x}_i)_{y_i} - {f_\theta}(\mathbf{x}_i)_{y_i}\bigg]\\
    &{s.t.} \quad \arg\!\max_y f_\vartheta(\mathbf{x}_f) \neq \arg\!\max_y f_\theta(\mathbf{x}_f)
\end{aligned}
\end{equation}

Consequently, this aggressive approach can significantly degrade the model's overall performance. Intuitively, a model can effectively forget an easily unlearnable training sample with minimal impact on its predictive performance. However, for a more challenging training sample, the required unlearning may exceed the acceptable tolerance, leading to a significant degradation in model performance. 

Model performance degradation can be quantified using two metrics: accuracy degradation and the increase in model loss. In our analysis, we evaluate both measures. Figure \ref{fig:acc_loss_test_svhn} presents the changes in accuracy and model loss on the test data following unlearning (results for the remaining data are provided in Appendix Section \ref{appendix:remain_loss_acc}). The results clearly demonstrate a strong positive correlation between the TPS and empirical unlearning outcomes, regardless of the unlearning algorithm employed. The observation highlights the effectiveness of TPS as a reliable index for measuring unlearning difficulty.

%It is worth noting that TPS can only be measured after performing the unlearning operation on a trained model. While it provides a valuable indication of unlearning difficulty, it cannot serve as a predictive index for forecasting unlearning outcomes in advance.

\begin{figure*}[t]
\centering
    \includegraphics[width=\textwidth]{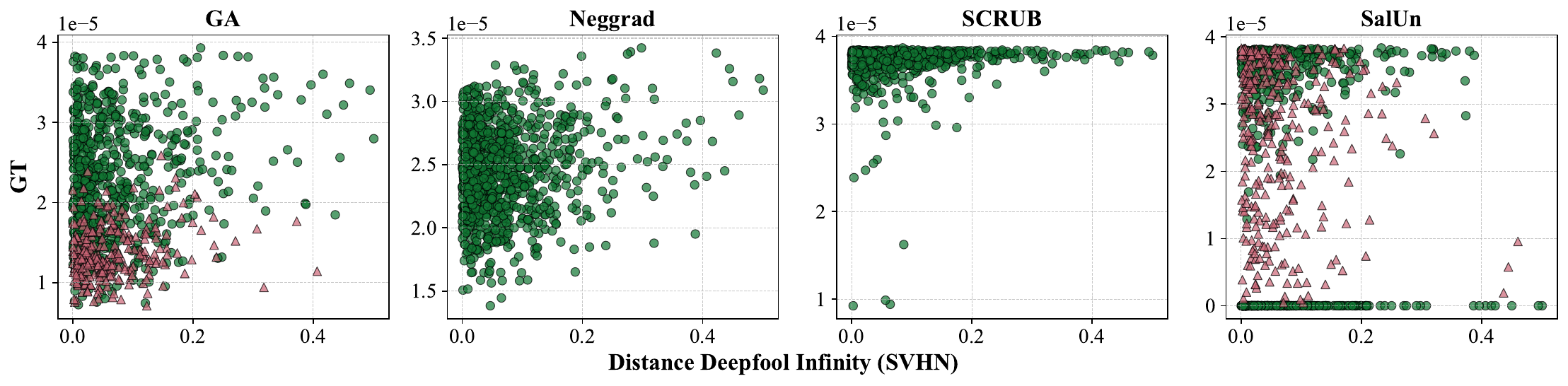}
    \includegraphics[width=\textwidth]{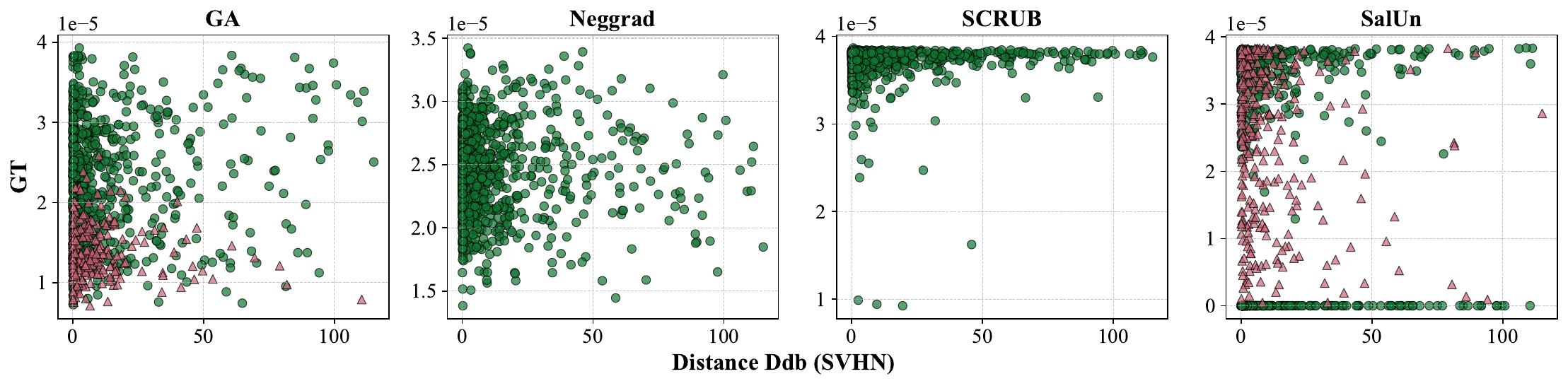}
   \vspace{-5mm}
        \caption{Effectiveness of using Geometric Distance to Decision Boundary (GDDB) as an index of unlearning difficulty. (Top) Distance to decision boundary estimated through DeepFool in adversarial learning literature. (Bottom) Distance to the decision boundary is estimated by treating the last layer of a neural network as a linear classifier. There is no observable correlation between empirical unlearning difficulty and training data's geometric distance to the decision boundary. The experiments is conducted on ResNet-18 model trained on SVHN dataset.}
    \label{fig:geomeric_distance}
\end{figure*}

\subsubsection{Distance of Parameter Shift}

Distance of Parameter Shift (DPS) comes from the same intuition of TPS but directly probes the layer-wise and activation-wise parameter shift distances. Such that seeking for a more precise estimation of unlearning difficulty~\cite{golatkar2020eternal, tarun2023fast}. Similar to TPS, the DPS can also denote the solution of an optimization task
\begin{equation}
\begin{aligned}
    &\min ||\vartheta - \theta||_1 \\
    &{s.t.} \quad \arg\!\max_y f_\vartheta(\mathbf{x}_f) \neq \arg\!\max_y f_\theta(\mathbf{x}_f)
\end{aligned}
\end{equation}

In particular, layer-wise distance measures the weight differences between the unlearned and original models, while activation-wise distance assesses their activation differences given the same input.

In our analysis, we assess the shift in model parameters during unlearning by examining the absolute difference. The results, presented in Figure \ref{fig:model_parm_shift_abs}, suggest that while parameter shift can serve as a reliable indicator of unlearning difficulty for general gradient-based unlearning approaches, it is not universally effective. Specifically, when an unlearning method targets specific neurons encoding unique properties of certain training data points, parameter shift may fail to accurately reflect unlearning difficulty. SalUn is a representative approach in such a category that leverages the weight saliency map, which violates the assumptions underlying DPS. In particular, an easily unlearnable sample may be distributed across many neurons through amortization, while a difficult-to-unlearn sample may be encoded by a small group of identifiable neurons. These identifiable neurons can be selectively modified or destroyed without significantly impacting the model's overall generalization, challenging the effectiveness of DPS as a universal measure of unlearning difficulty. 

%Similar to TPS, the DPS can only be measured after performing the unlearning operation on a trained model, limiting it to be a valid index of unlearning difficulty.

% {\color{red} Wuga: Stopped my editing here given we are still waiting for some experimental results to continue the writing. Time mark: 2025-01-28 11PM}
% {\color{blue} Mahtab: I will try to follow your footsteps for the section. Hopefully it will be helpful}

\subsubsection{Geometric Distance to Decision Boundary}
A previous study by \cite{chen2023boundary} found that unlearning by retraining the model pushes the forget samples to move around the border of other clusters. As a result, samples located near cluster boundaries in the decision space are more likely to be predicted with high uncertainty. Based on this observation, it is natural to consider a data point's distance to the decision boundary as a potential index for quantifying unlearning difficulty.

While the hypothesis above is theoretically compelling, it is often challenging to verify given the complexity of modern machine learning models, where the distance between a data point and the decision boundary are hard to measure. We, therefore, conducted our analysis by approximating the distance through DeepFool \citep{moosavi2016deepfool} and linear model approximation. Formally, for DeepFool, we define the data point's distance to the decision boundary through the norm of an $\epsilon$-ball
\begin{equation}
\begin{aligned}
    &\min ||\epsilon||\\
    &{s.t.} \quad \arg\!\max_y f_\theta(\mathbf{x}_f+\mathbf{\epsilon}) \neq \arg\!\max_y f_\theta(\mathbf{x}_f),
\end{aligned}
\end{equation}
where there is no unlearned model $f_\vartheta$ involved in the index. Alternatively, we can treat the last layer of a complex neural network as a linear model defined on the learned representations $\phi(\mathbf{x})$ such that the distance of a data point to decision boundary can be easily estimated through vector projection 
\begin{equation}
\begin{aligned}
    &\min_{\mathbf{w}_k} \frac{\mathbf{w}_k^\top \phi(\mathbf{x}_f) +b_k}{||\mathbf{w}_k||} \quad \forall (\mathbf{w}_k, b_k) \in f_\theta^L,
\end{aligned}
\end{equation}
where $f_\theta^L$ denotes the last layer of a trained ML model.

\begin{figure*}[t]
\centering
    \includegraphics[width=\textwidth]{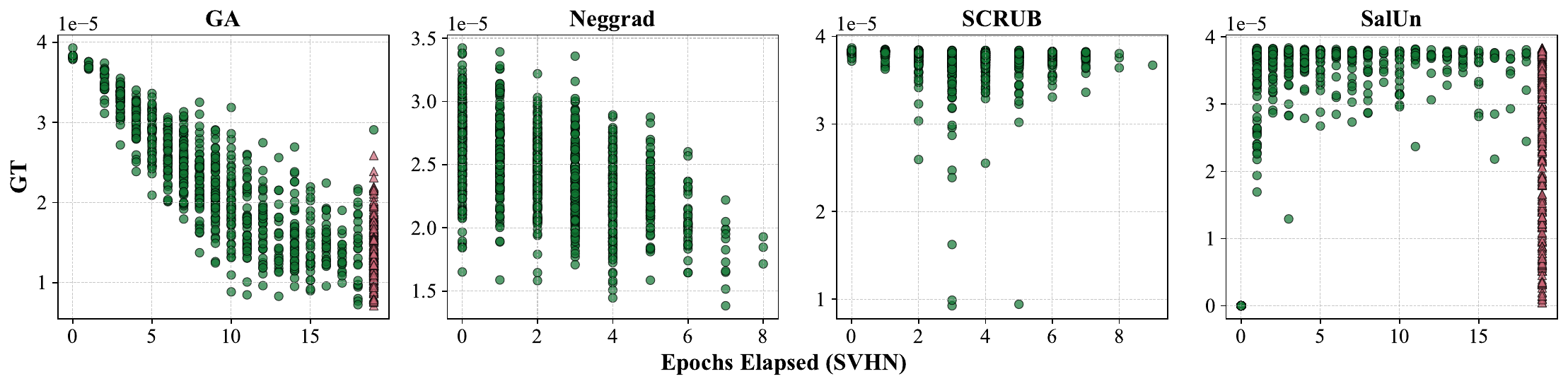}
   \vspace{-5mm}
        \caption{Effectiveness of using Number of Unlearning Epochs (NUE) as an index of unlearning difficulty. 
        %(Top) Number of epochs used to achieve guaranteed unlearning (flip of decision). (Bottom) Wall clock time elapsed to achive guaranteed unlearning. 
        We observed noisy negative alignment between NUE and GT for gradient based approaches. For SCRUB and SalUn, there is no observable correlation. The experiments are conducted on ResNet-18 model trained on SVHN dataset.}
    \label{fig:unlearning_steps}
\end{figure*}

Figure~\ref{fig:geomeric_distance} presents our experimental results. Unfortunately, we do not observe a clear correlation between GDDB and unlearning difficulty when removing a single data point from the model, regardless of the unlearning algorithm used. While there may be concerns regarding the accuracy of our proposed distance approximation, our overall assessment suggests that GDDB is not a reliable index for measuring unlearning difficulty, at least in the context of single data removal tasks.

Several studies~\cite{cotogni2023duck, chen2024machine, foster2024} have explored the idea of manipulating the position of the decision boundary to facilitate machine unlearning, based on a hypothesis similar to the one we presented. Interestingly, these studies report promising unlearning performance, which appears to contradict our observations. However, upon closer examination of their experimental setups, we find that decision boundary-based unlearning algorithms are typically applied to the removal of an entire class or cluster of data points, rather than individual data points. This raises concerns about the practical applicability of such approaches, particularly in enforcing Right to be Forgotten regulations, which often necessitate the removal of specific individual data points.

\subsubsection{Number of Unlearning Steps}
Number of Unlearning Steps (NUS) evaluates the computational efficiency of the unlearning operations, indicating how quickly the model can be updated to forget specified data. For a given unlearning algorithm, the metric can be approximated through wall clock duration \cite{nguyen2022survey} or the number of unlearning epochs. As the wall clock time depends on the hardware configuration of system (computational load on the system), we also consider the number of unlearning epochs associated with each algorithm. From the experimental results provided in Figure~\ref{fig:unlearning_steps}, we observe that the increasing of unlearning epochs is positively correlated with the unlearning difficulty. The easier samples can be unlearned with smaller unlearning epochs. SalUn continues unlearning process for difficult samples until the reaching the final epoch. Often these data points cannot be fully unlearned by the end of the process that indicates such samples were difficult and unlearning them was not successful.

\begin{figure*}[t]
    \centering
    \includegraphics[width=\textwidth]{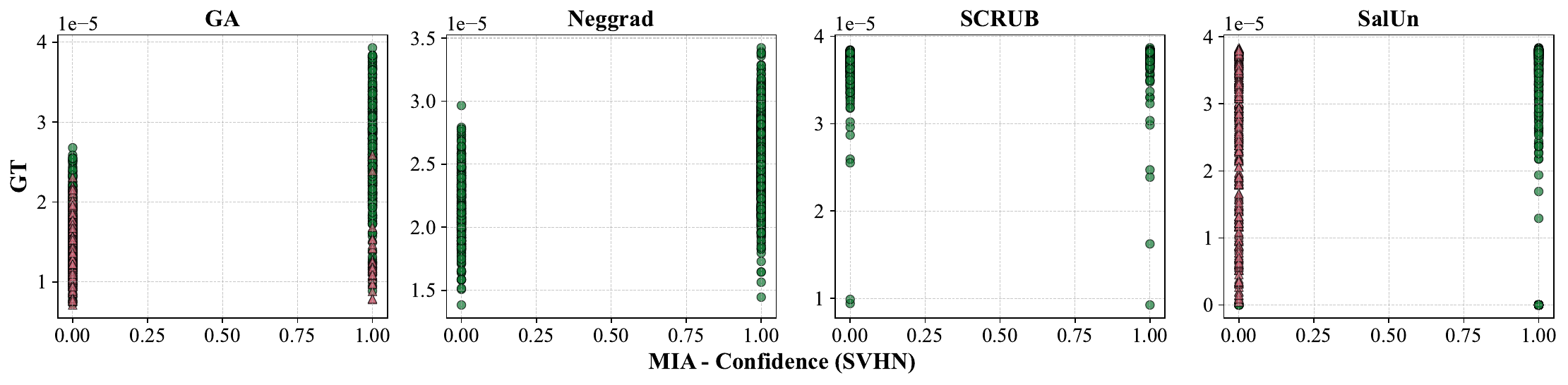}
    \vspace{-5mm}
    \caption{Effectiveness of MIA as index of unlearning difficulty. A prediction of "1" indicates successful unlearning, where the unlearned model no longer retains information about the data, while "0" signifies failed unlearning, where the model still remembers the unlearned samples.  We observed noisy positive alignment between MIA and GT for gradient based approaches. The larger the GT, the easier the data point for unlearning and this is positively associated with MIA="1". For SCRUB there is no observable correlation. The experiments are conducted on ResNet-18 model trained on SVHN dataset.}
    \label{fig:mia}
\end{figure*}

\subsubsection{Resistance to Membership Inference Attack}

In Membership Inference Attacks (MIA)\cite{chen2021machine, golatkar2021mixed, song2019privacy}, adversaries exploit the model’s outputs, such as confidence scores, to infer whether a specific data point was part of the training set, without requiring direct access to the model’s internal parameters. Within the context of model unlearning (MU), this metric is employed to detect residual imprints of the forgetting set $D_f$ in the unlearned model\cite{chen2021machine}. Specifically, an MIA is conducted on the unlearned model (\textit{MIA-Correctness}) to evaluate the extent to which data points from $D_f$ are correctly classified as non-training samples, serving as an indicator of successful unlearning.

The effectiveness of MIA is evaluated by the proportion of samples identified as "forgotten" (True Negatives, \textit{TN}) relative to the total size of the forgetting set \( |D_f| \). Ideally, after unlearning, the model \( \theta_u \) should have successfully "forgotten" the information associated with the samples in the forgetting set. To assess MIA efficacy, we employed a confidence-based attack method~\cite{song2019privacy}. 

When $ |D_f| = 1 $, the inference attack simplifies to binary classification: a prediction of "1" indicates successful unlearning, where the unlearned model no longer retains information about the data, while "0" signifies failed unlearning, where the model still remembers the unlearned samples. The results, presented in Figure \ref{fig:mia}, show a positive correlation with ground truth (GT). Easier samples, characterized by higher GT values, tend to be unlearned more effectively, whereas more challenging samples, associated with lower GT values, are less successfully unlearned.

\begin{figure*}[t]
    \centering
    \includegraphics[width=\textwidth]{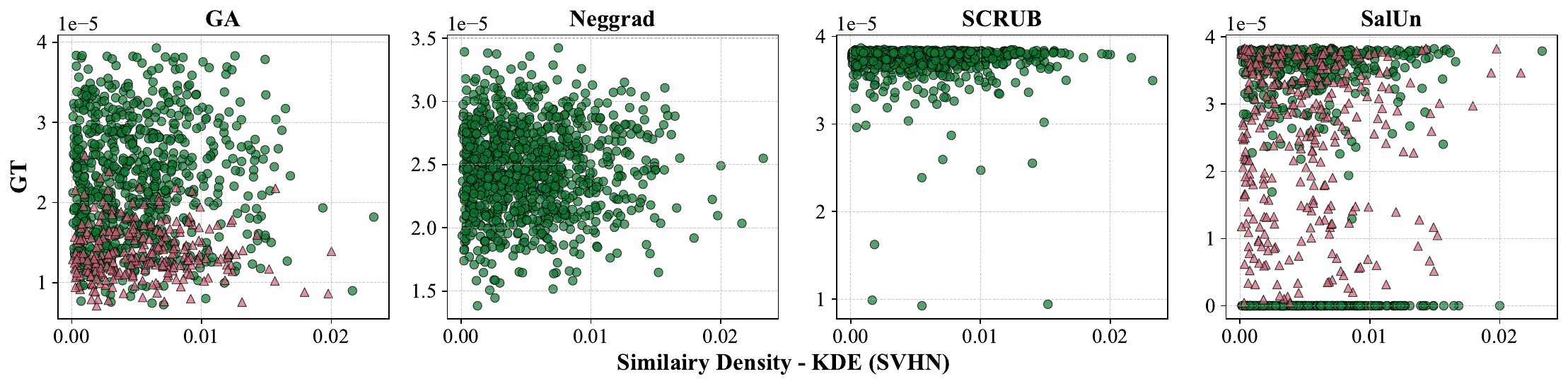}
    \vspace{-5mm}
    \caption{Effectiveness of Size of Unlearning Expansion (SUE) as an index of unlearning difficulty. The similarity each data point is measured by KDE at that point. Higher density values indicate that the data point lies in a region with many similar samples. From the obtained results there is no correlation between SUE and GT. The experiments are conducted on ResNet-18 model trained on SVHN dataset.}
    \label{fig:size}
\end{figure*}

\subsubsection{Size of Unlearning Expansion}
Altering prediction outcome of target sample may negatively impact model prediction on similar samples. When a guaranteed unlearning is desired, one might need to expand unlearning operation to a broader training sample set (the similar data samples) such that unlearning of target sample with respect to decision shift can be successful \cite{chen2023boundary}. The ML models process the datapoints collectively and learn the pattern from the whole data \cite{xu2024machine}, therefore it can be perceived that to unlearn a single sample the collective pattern \cite{schelter2021hedgecut} should be disrupted. Regarding the unlearning algorithm, the requirement to expand the size of the forget set, referred to as the Size of Unlearning Expansion (SUE), is crucial for ensuring effective unlearning. We hypothesize that a datapoint with a larger and stronger SUE should be more challenging to unlearn. To test this hypothesis, we compute the similarity between each unlearning sample $\mathbf{x}_f$ and the remaining training data using Kernel Density Estimation (KDE), as follows:
\begin{equation}
P(\mathbf{x}_f) = \frac{1}{|D_r|} \sum_{i=1}^{n} \psi\left(\mathbf{x}_f, \mathbf{x}_i\right)/t
\end{equation}
where the $\psi$ is a kernel function, and $t$ a smoothing parameter (temperature). The evaluation results are presented in Figure \ref{fig:size}, which show no clear relationship between SUE and unlearning difficulty. These findings suggest that unlearning algorithms should focus on the forget set, rather than relying solely on the remaining data to achieve unlearning. Even for the NegGrad method, which fine-tunes the model on $D_r$, the SUE does not appear to have a significant effect. This indicates that relying solely on the remaining data for forgetting does not guarantee effective unlearning. To achieve optimal results, it is essential to process both $D_r$ and $D_f$.

\subsubsection{Quantitative Evaluation}
Further, we employed the Spearman Correlation metric to assess the strength and direction of correlation between the GT and discussed factors. This metric is a non-parametric measure of the strength and direction of the monotonic relationship between two variables. It evaluates how well the relationship between two variables can be described by a monotonic function, whether linear or not. The results of the Spearman Correlation analysis are shown in Table \ref{tab:metric_comparison}, where we report the correlation coefficients (Corr) and the associated p-values for each factor and unlearning method. GA, NegGrad, and SCRUB largely exhibited the expected results, showing a strong positive correlation between test accuracy and GT, and strong negative correlations between GT and factors such as Model Distance, elapsed unlearning time, and unlearning epochs. SalUn on SVHN and MNIST demonstrated positive correlations with test accuracy, Model Distance (KL), and Forget Predictive Probability Difference, indicating its effectiveness in unlearning samples with high predictive probability differences. For CIFAR-10, SalUn performance shows negative correlations, particularly with test accuracy and Forget Loss Difference which suggests the SalUn may struggle with certain difficult samples, especially when unlearning is more challenging. 
\begin{table*}[h]
    \centering
    \scriptsize
    \resizebox{\textwidth}{!}{%
    \begin{tabular}{c|l|cc|cc|cc|cc}
        \toprule
        && \multicolumn{2}{c|}{SalUn} & \multicolumn{2}{c|}{NG} & \multicolumn{2}{c}{SCRUB} & \multicolumn{2}{c}{GA} \\
        & Factor/Metric & Corr & p-value & Corr & p-value & Corr & p-value & Corr & p-value \\
        \midrule
        \multirow{9}{*}{\rotatebox[origin=c]{90}{{SVHN}}} &Test Accuracy (TPS) & 0.2507 & $8.39 \times 10^{-16}$ & 0.9981 & 0.0 & 0.9879 & 0.0 & 0.9996 & 0.0 \\ 
        &Model Distance - KL (DPS) & 0.4414 & $6.29 \times 10^{-49}$ & -0.5548 & $8.84 \times 10^{-82}$ & -0.7820 & $4.09 \times 10^{-207}$ & -0.7988 & $1.66 \times 10^{-222}$ \\
        &Model Distance - Absolute (DPS) & 0.3648 & $7.68 \times 10^{-33}$ & -0.5706 & $1.89 \times 10^{-87}$ & -0.8133 & $6.36 \times 10^{-237}$ & -0.7898   & $3.92 \times 10^{-214}$ \\
        &Forget Predictive Probability Difference & 0.7974 & $4.06 \times 10^{-221}$ & -0.1288 & $4.42 \times 10^{-5}$ & -0.6125 & $5.01 \times 10^{-104}$ & 0.3618 & $2.76 \times 10^{-32}$ \\
        &Test Predictive Probability Difference & -0.2547 & $2.87 \times 10^{-16}$ & -0.9999 & 0.0 & -0.9999 & 0.0  & -0.9998 & 0.0 \\
        &Forget Loss Difference & -0.2347 & $5.56 \times 10^{-14}$ & 0.5691 & $6.85 \times 10^{-87}$ & -0.6226 & $2.05 \times 10^{-108}$ & -0.7194 & $4.22 \times 10^{-160}$  \\
        &Test loss Difference & 0.2507 & $8.56 \times 10^{-16}$ & 0.8467 & $1.09 \times 10^{-275}$ & 0.9796 & 0.0 & -0.7899 & $3.92 \times 10^{-214}$  \\
        &Elapsed Time  (NUE) & 0.4245 & $5.24 \times 10^{-45}$ & -0.5339 & $8.98 \times 10^{-75}$ & -0.3442 & $3.37 \times 10^{-29}$ & -0.5472 & $3.58 \times 10^{-79}$ \\
        &Epochs Elapsed  (NUE) & 0.4360 & $1.20 \times 10^{-47}$ & -0.5400 & $8.88 \times 10^{-77}$ & -0.3571 & $1.95 \times 10^{-31}$ & -0.8477 & $5.57 \times 10^{-277}$ \\
        \midrule
        \multirow{9}{*}{\rotatebox[origin=c]{90}{MNIST}} 
        & Test Accuracy (TPS) & 0.4385 & $3.04 \times 10^{-48}$ & 0.9871 & 0.0 & 0.9823 & 0.0 & 0.9930 & 0.0 \\
        & Model Distance - KL (DPS) & 0.4502 & $4.58 \times 10^{-51}$ & -0.4739 & $4.07 \times 10^{-57}$ & -0.6461 & $3.25 \times 10^{-119}$ & -0.4302 & $2.56 \times 10^{-46}$ \\
        & Model Distance - Absolute (DPS) & 0.2633 & $2.52 \times 10^{-17}$ & -0.7123 & $1.33 \times 10^{-155}$ & -0.6376 & $3.47 \times 10^{-115}$ & -0.7201 & $1.39 \times 10^{-160}$ \\
        & Forget Predictive Probability Difference & 0.6231 & $1.23 \times 10^{-108}$ & -0.4404 & $1.08 \times 10^{-48}$ & -0.3985 & $2.11 \times 10^{-39}$ & -0.4310 & $1.74 \times 10^{-46}$ \\
        & Test Predictive Probability Difference & -0.9332 & 0.0 & -0.9999 & 0.0 & -0.9901 & 0.0 & -0.9940 & 0.0 \\
        & Forget Loss Difference & -0.2113 & $1.48 \times 10^{-11}$ & 0.6480 & $3.71 \times 10^{-120}$ & -0.6430 & $1.00 \times 10^{-117}$ & -0.7052 & $3.36 \times 10^{-151}$ \\
        & Test Loss Difference & 0.9336 & 0.0 & 0.9074 & 0.0 & 0.9489 & 0.0 & 0.9295 & 0.0 \\
        & Elapsed Time  (NUE) & 0.0990 & $1.72 \times 10^{-3}$ & -0.6504 & $2.69 \times 10^{-121}$ & -0.3930 & $2.86 \times 10^{-38}$ & -0.4510 & $2.89 \times 10^{-51}$ \\
        & Epochs Elapsed  (NUE) & 0.1076 & $6.56 \times 10^{-4}$ & -0.7093 & $9.06 \times 10^{-154}$ & -0.3983 & $2.28 \times 10^{-39}$ & -0.5050 & $7.77 \times 10^{-66}$ \\  
        \midrule
        \multirow{9}{*}{\rotatebox[origin=c]{90}{CIFAR 10   }} 
        & Test Accuracy (TPS) & -0.3263 & $3.09 \times 10^{-26}$ & 0.9769 & 0.0 & 0.9144 & 0.0 & 0.9963 & 0.0 \\
        & Model Distance - KL (DPS) & -0.1315 & $3.02 \times 10^{-5}$ & -0.4264 & $1.94 \times 10^{-45}$ & -0.7627 & $4.20 \times 10^{-191}$ & -0.1106 & $4.58 \times 10^{-4}$ \\
        & Model Distance - Absolute (DPS) & -0.1753 & $2.40 \times 10^{-8}$ & -0.4315 & $1.31 \times 10^{-46}$ & -0.7730 & $2.02 \times 10^{-199}$ & -0.6060 & $2.64 \times 10^{-101}$ \\
        & Forget Predictive Probability Difference & 0.7503 & $1.41 \times 10^{-181}$ & 0.4613 & $7.70 \times 10^{-54}$ & -0.1087 & $5.73 \times 10^{-4}$ & 0.2451 & $3.77 \times 10^{-15}$ \\
        & Test Predictive Probability Difference & 0.3405 & $1.46 \times 10^{-28}$ & -0.9781 & 0.0 & -0.9964 & 0.0 & -0.9966 & 0.0 \\
        & Forget Loss Difference & -0.4466 & $3.55 \times 10^{-50}$ & 0.4234 & $9.16 \times 10^{-45}$ & -0.7780 & $1.04 \times 10^{-203}$ & -0.5810 & $2.54 \times 10^{-91}$ \\
        & Test Loss Difference & -0.3197 & $3.37 \times 10^{-25}$ & 0.9266 & 0.0 & 0.7244 & $2.08 \times 10^{-163}$ & 0.9363 & 0.0 \\
        & Elapsed Time  (NUE) & -0.1961 & $3.96 \times 10^{-10}$ & -0.4084 & $1.72 \times 10^{-41}$ & -0.6218 & $4.95 \times 10^{-108}$ & -0.5445 & $2.89 \times 10^{-78}$ \\
        & Epochs Elapsed  (NUE) & -0.1974 & $3.06 \times 10^{-10}$ & -0.4096 & $9.55 \times 10^{-42}$ & -0.6238 & $5.97 \times 10^{-109}$ & -0.5788 & $1.74 \times 10^{-90}$ \\
        \bottomrule
    \end{tabular}}
    \caption{The correlation and p-value results between the GT score and various factors examined in this study for different unlearning methods. The term "metrics" following the factor abbreviations indicates the specific measurements used in the empirical evaluation.}
    \label{tab:metric_comparison}
\end{table*}

\subsubsection{Further Discussion}

One of the interesting observation from our experiments has been the relation of SUE and unlearning difficulty. Despite SUE being strongly associated with model generalization, we found no significant correlation with unlearning difficulty. This could be due to unlearning algorithms specifically targeting the forget set, which may reduce SUE's impact. Also, Unlearning difficulty can be assessed in two ways: by evaluating model damage (e.g., decreased accuracy, increased error, parameter shifts, and processing time), or by assessing an algorithm’s ability to unlearn specific samples. GA struggled to unlearn the most difficult samples, significantly harming test accuracy. In contrast, while NegGrad and SCRUB successfully unlearned all samples, challenging samples still hurt model performance. Our analysis suggests that unlearning difficulty is best reflected by post-unlearning model performance. However, future research should explore how to predict unlearning difficulty based on model characteristics before the unlearning process begins.
% The investigation and comparison of unlearning difficulty can be approached from two perspectives. First, it can be evaluated based on the damage caused to the model, such as lower test accuracy, increased error on test data, significant shifts in model parameters, and longer processing times. Second, it can be assessed through the inability of an algorithm to effectively unlearn certain samples. GA demonstrated a limited ability to unlearn the most challenging samples; moreover, the process of unlearning these difficult samples led to a significant reduction in the model's predictive performance on test data. In contrast, Although NegGrad and SCRUB were able to successfully unlearn all samples, the presence of difficult samples significantly impaired the model’s accuracy and increased its loss on test data, highlighting the damaging impact unlearning such samples on performance. From our assessment Unlearning difficulty is best reflected by the performance of the model after unlearning. However, developing a criterion to link model characteristics and difficulty factors before the initiation of unlearning process if future research direction that peruse. 

\section{Alternate View}
% While this paper advocates for analyzing machine unlearning through the lens of instance-level difficulty, there are alternative perspectives that needs to be discussed. A significant counterargument holds that instance-level analysis may be misguided, as it fails to capture the reality of how neural networks learn and store information. Modern deep learning models encode large amount of information which making the "difficulty" of unlearning individual samples unrealistic. Their learned representations are entangled, and complex which attempting to isolate and quantify the unlearning difficulty of single instances may be fundamentally flawed. Also the "right to be forgotten" rarely involves removing individual data points. Instead, organizations typically need to handle batch deletion requests or remove entire categories of data. From this perspective, focusing on instance-level unlearning difficulty may be solving the wrong problem - the real challenge lies in developing efficient batch unlearning techniques that maintain model utility while removing broader categories of information. Additionally machine learning models learn patterns rather than memorizing individual examples, hence concept of removing a single sample's influence could be unrealistic expectation. While instance-level unlearning difficulty provides valuable insights, these opposing viewpoints highlight important limitations and alternative directions that deserve careful consideration from the machine unlearning community. 
The study by Zhao et al. \cite{zhao2024makes} assessed unlearning difficulty using data space entanglement between the remaining and forget sets and the model's memorization strength, with a proxy metric based on model accuracy across these sets. However, our analysis suggests that accuracy on the remaining set may be less influenced by unlearning than test set accuracy, as the remaining set is directly optimized during the unlearning process. Zhao et al. defined unlearning difficulty using two factors: Data Space Entanglement (ES), which compares the mean embeddings of the retain and forget sets, and Memorization Strength, which examines the model's predicted probabilities before and after unlearning. We argue that averaging embeddings may fail to capture data space entanglements accurately, especially in models that have not learned the data space well. Furthermore, memorization strength is influenced by the unlearning algorithm used. In this paper, we investigate unlearning difficulty independently of specific unlearning algorithms, examining it from both data and model perspectives.

\section{Conclusion}

In this paper, we examined the difficulty of machine unlearning in response to the increasing demand for this process.  To facilitate the analysis of unlearning difficulty, we quantify the difficulty of unlearning operation for individual data through a harmonic average of unlearned model performance shift. 
We summarized the six unlearning feasibility factors that are commonly assumed to be effective on assessing the difficulty of unlearning, including as size of unlearning expansion, tolerance of performance shift, resistance to membership attack, etc.  Our empirical evaluation shows that four out of six factors examined can provide guidance on correctly identifying easy and difficult samples, which shows the need of understanding the difficulty of machine unlearning.

\bibliography{example_paper}
\bibliographystyle{icml2025}

%%%%%%%%%%%%%%%%%%%%%%%%%%%%%%%%%%%%%%%%%%%%%%%%%%%%%%%%%%%%%%%%%%%%%%%%%%%%%%%
%%%%%%%%%%%%%%%%%%%%%%%%%%%%%%%%%%%%%%%%%%%%%%%%%%%%%%%%%%%%%%%%%%%%%%%%%%%%%%%
% APPENDIX
%%%%%%%%%%%%%%%%%%%%%%%%%%%%%%%%%%%%%%%%%%%%%%%%%%%%%%%%%%%%%%%%%%%%%%%%%%%%%%%
%%%%%%%%%%%%%%%%%%%%%%%%%%%%%%%%%%%%%%%%%%%%%%%%%%%%%%%%%%%%%%%%%%%%%%%%%%%%%%%
\newpage
\appendix
\onecolumn
\section{Impact Statement}
This paper highlights a critical gap in machine unlearning (MU) research by emphasizing the variability in unlearning difficulty at the instance level, challenging the assumption that unlearning is uniformly effective for all data points. Through empirical analysis of unlearning outcomes across multiple algorithms and datasets, we identify key factors that influence unlearning difficulty, offering a deeper understanding of the complexities involved. These findings not only call for more refined evaluation metrics that account for instance-level variability but also open up new avenues for developing predictive tools to improve MU effectiveness in real-world applications, reducing computational costs and enhancing the practical usability of unlearning methods.

\section{Unlearning Algorithms}

\begin{table}[h]
\centering
\begin{tabular}{c|ccc}
\hline
Algorith             & Unlearning rate & Unlearning steps &  \\ \hline
Gradient Ascent (GA) & 1e-4              & 20                               \\
NegGrad             & 1e-4               & 20                               \\
SCRUB                & 1e-3               & 20                               \\
SalUn                & 0.01               & 20                               \\
Fine Tune (FT)       & 1e-4               & 20                            \\ \hline
\end{tabular}
\label{tab:unlearn_algo}
\end{table}

\section{Datasets and Models}
\label{appendix:data_param}
The summary of each model, training parameters and dataset associated for to that model is given in this table. 

\begin{table*}[ht!]
\centering
\caption[Summary of models and parameters used for evaluation of unlearning algorithm across three datasets.]{This table details the datasets and models used in evaluating unlearning algorithms, specifying the models applied to each dataset, including the number of layers, batch sizes, number of classes, learning rates, and sample sizes. The information presented provides insight into the computational frameworks employed to analyze MNIST, CIFAR-10, and SVHN datasets, demonstrating the diversity of approaches used in the study.}
\resizebox{\textwidth}{!}{%
\begin{tabular}{@{}lcccccc@{}}
\toprule
Dataset         & Model            & Layers & Batch Size & Number of Classes  & Learning Rate & Samples \\ \midrule
MNIST           & ResNet18         & 18     & 150        & 10                 & 0.001         & 54000   \\
CIFAR10         & ResNet18         & 18     & 150        & 10                 & 0.01          & 45000   \\ 
SVHN            & ResNet18         & 18     & 64         & 10                 & 0.001         & 58000   \\ \bottomrule

\end{tabular}%
}
\label{tab:data}
\end{table*}

\section{Remaining data performance shift}
\label{appendix:remain_loss_acc}
    \begin{figure*}[h!]
    \centering
        \includegraphics[width=\textwidth]{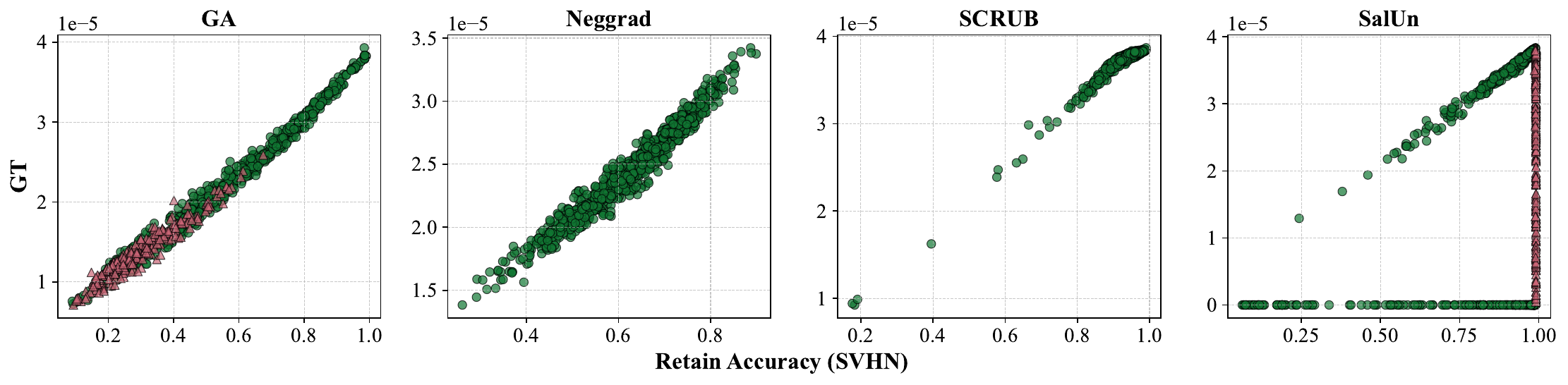}
        \includegraphics[width=\textwidth]{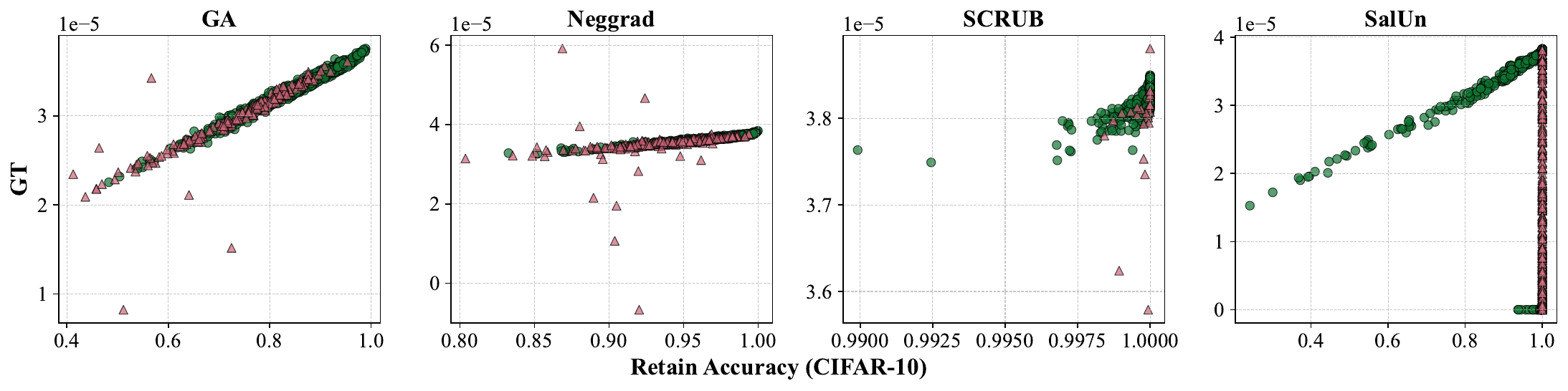}
        \includegraphics[width=\textwidth]{icml2025/figs/single_unlearn_figures/train_accruacy_cifar.pdf}
        \caption{Accuracy of the remaining data (\(D_r\)) post-unlearning, presented consecutively from top to bottom for the SVHN, CIFAR-10, and MNIST datasets. For the four different unlearning algorithms, we note there is a consistent positive alignment between accuracy of remaining data and empirical unlearning outcome (GT).}
        \label{fig:remain_acc}
        \vspace{-0.4cm}
    \end{figure*}

\newpage

\section{Distance of Parameter Shift}
\label{appendix:param_shift}

\begin{figure*}[h!]
\centering
    \includegraphics[width=\textwidth]{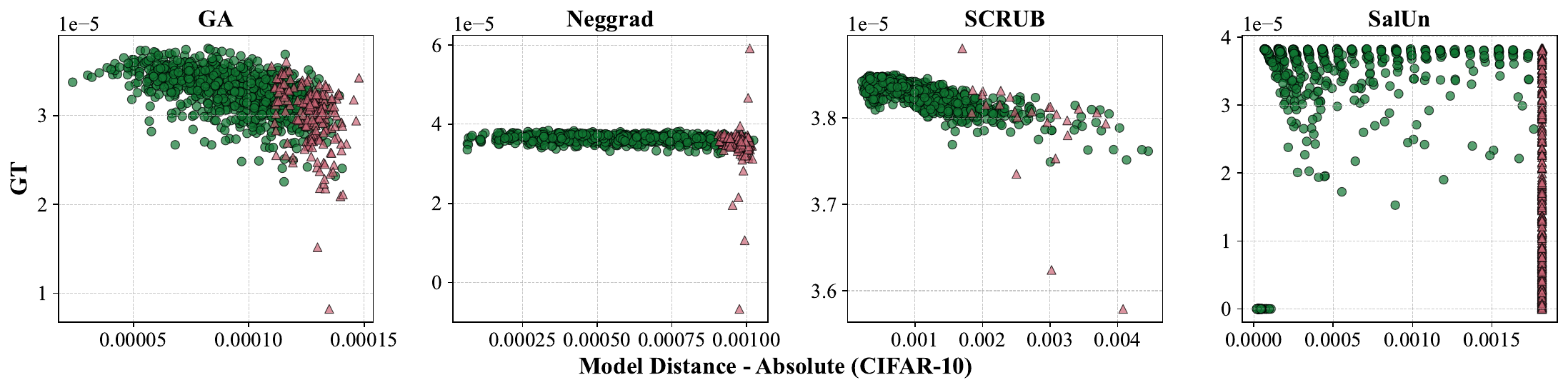}
    \includegraphics[width=\textwidth]{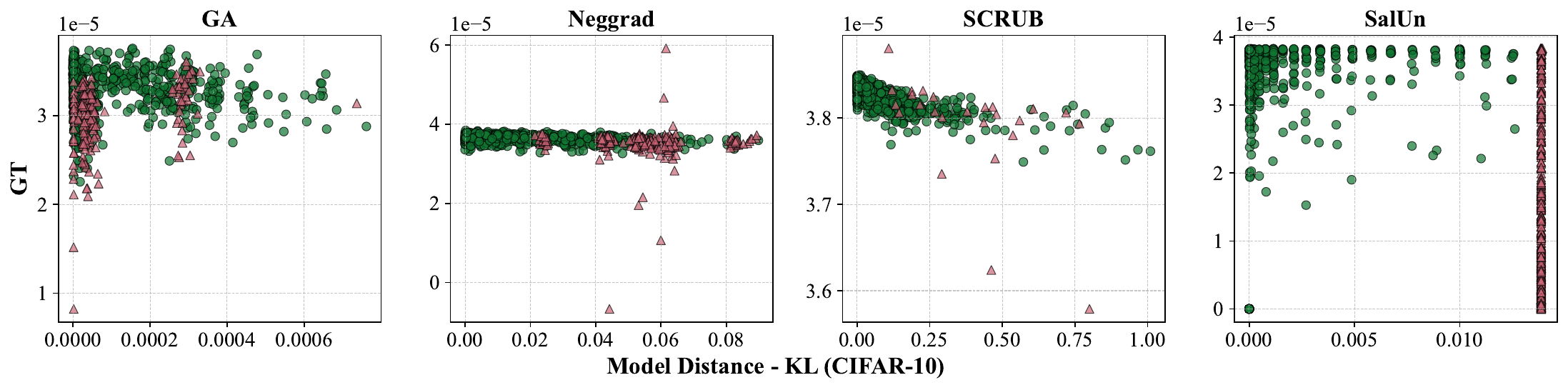}
   \vspace{-5mm}
        \caption{Effectiveness of using Distance of Preference Shift (DPS) as index of unlearning difficulty. (Top) Absolute average layer-wise distance of the model's weights pre- and post-unlearning. (Bottom) KL-divergence of model's parameters before and after unlearning. The experiments is conducted on ResNet-18 model trained on CIFAR10 dataset.}
    \label{fig:model_parm_shift_abs_cifar}
\end{figure*}

\begin{figure*}[h!]
\centering
    \includegraphics[width=\textwidth]{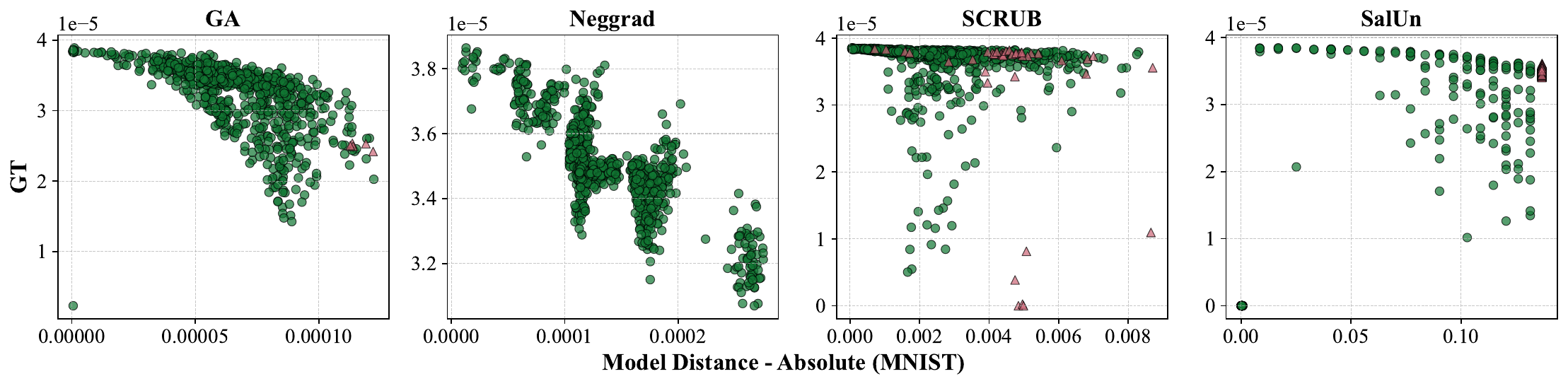}
    \includegraphics[width=\textwidth]{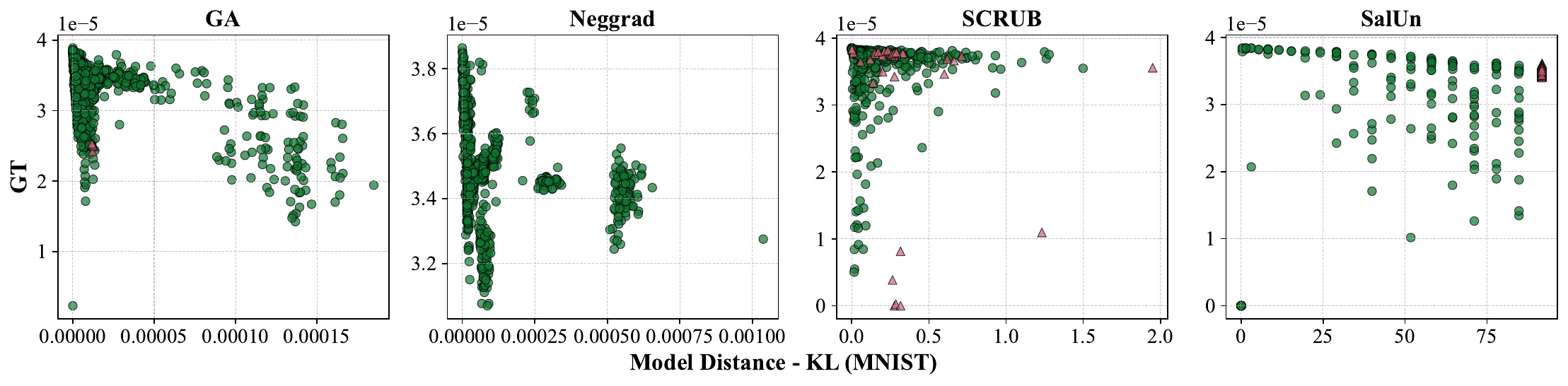}
   \vspace{-5mm}
        \caption{Effectiveness of using Distance of Preference Shift (DPS) as index of unlearning difficulty. (Top) Absolute average layer-wise distance of the model's weights pre- and post-unlearning. (Bottom) KL-divergence of model's parameters before and after unlearning.} The experiments is conducted on ResNet-18 model trained on MNIST dataset.
    \label{fig:model_parm_shift_abs_mnist}
\end{figure*}

\newpage

\section{Geometric Distance to Decision Boundary}

\begin{figure*}[h!]
\centering
    \includegraphics[width=\textwidth]{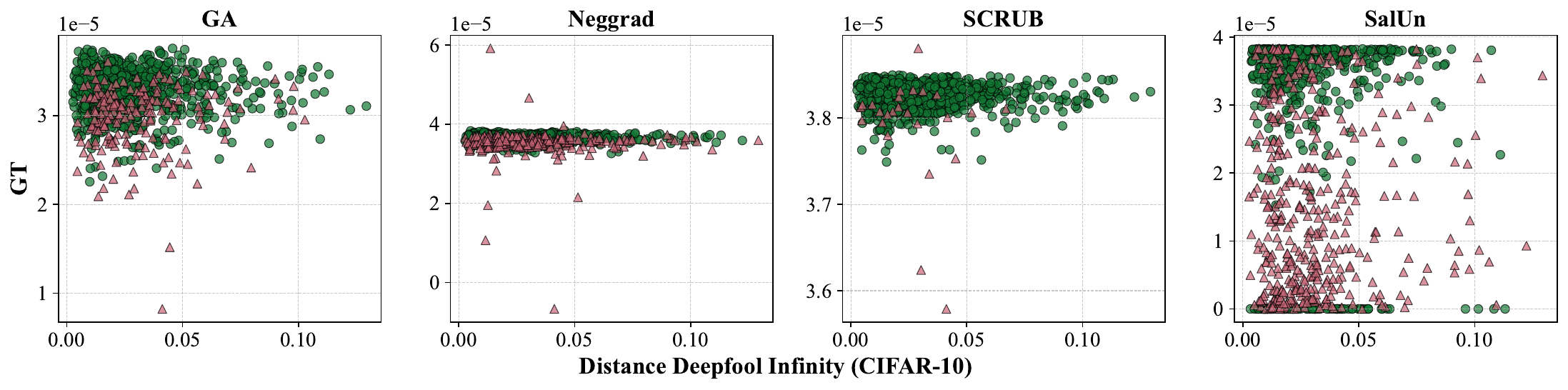}
    \includegraphics[width=\textwidth]{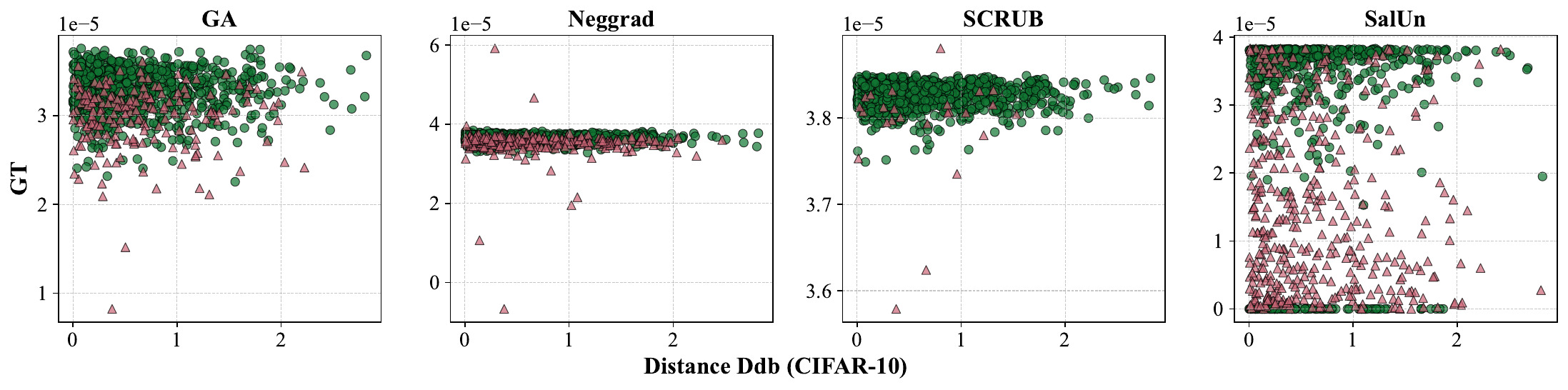}
   \vspace{-5mm}
        \caption{Effectiveness of using Geometric Distance to Decision Boundary (GDDB) as index of unlearning difficulty. (Top) Distance to decision boundary estimated through DeepFool in adversarial learning literature. (Bottom) Distance to decision boundary estimated by treating the last layer of neural network as linear classifier. The experiments is conducted on ResNet-18 model trained on CIFAR10 dataset.}
    \label{fig:geomeric_distance_cifar}
\end{figure*}

\begin{figure*}[h!]
\centering
    \includegraphics[width=\textwidth]{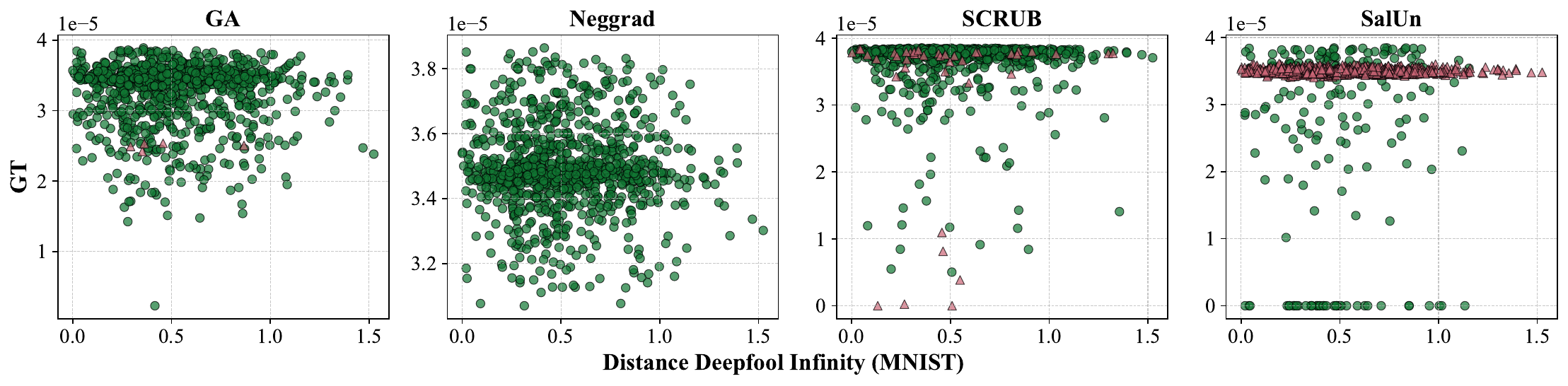}
    \includegraphics[width=\textwidth]{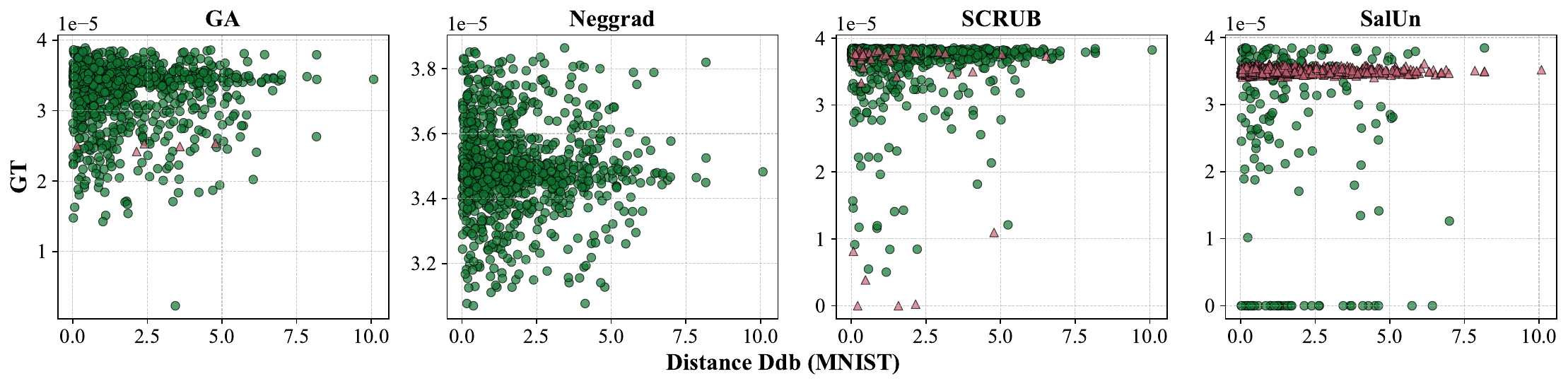}
   \vspace{-5mm}
        \caption{Effectiveness of using Geometric Distance to Decision Boundary (GDDB) as index of unlearning difficulty. (Top) Distance to decision boundary estimated through DeepFool in adversarial learning literature. (Bottom) Distance to decision boundary estimated by treating the last layer of neural network as linear classifier. The experiments is conducted on ResNet-18 model trained on MNIST dataset.}
    \label{fig:geomeric_distance_mnist}
\end{figure*}

\newpage

\section{Number of Unlearning Epochs}
\begin{figure*}[h!]
\centering
    \includegraphics[width=\textwidth]{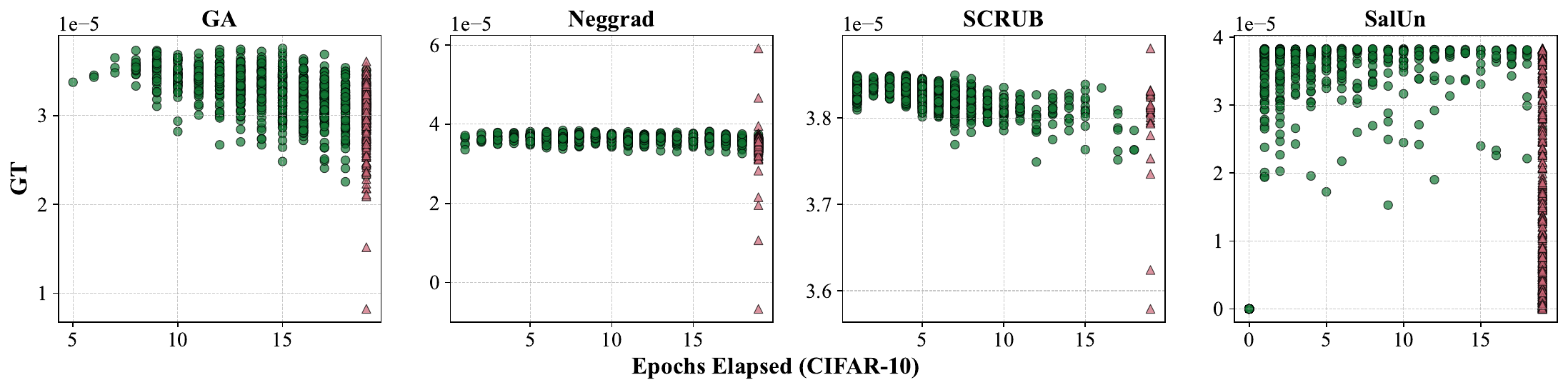}
    \includegraphics[width=\textwidth]{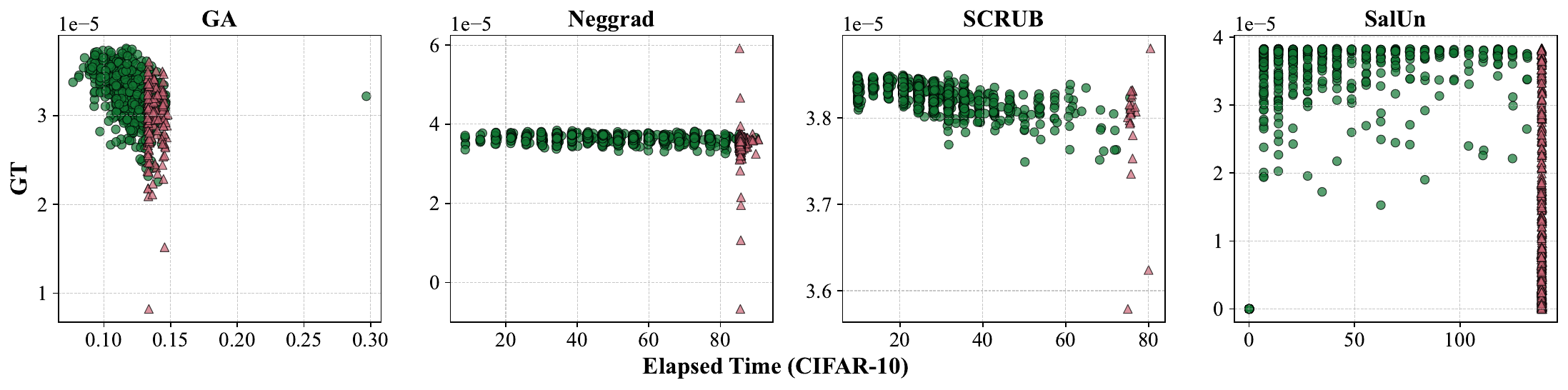}
   \vspace{-5mm}
        \caption{Effectiveness of using Number of Unlearning Epochs (NUE) as index of unlearning difficulty. (Top) Number of epochs used to achieve guaranteed unlearning (flip of decision). (Bottom) Wall clock time elapsed to achive guaranteed unlearning. The experiments is conducted on ResNet-18 model trained on CIFAR10 dataset.}
    \label{fig:unlearning_steps_cifar}
\end{figure*}

\begin{figure*}[h!]
\centering
    \includegraphics[width=\textwidth]{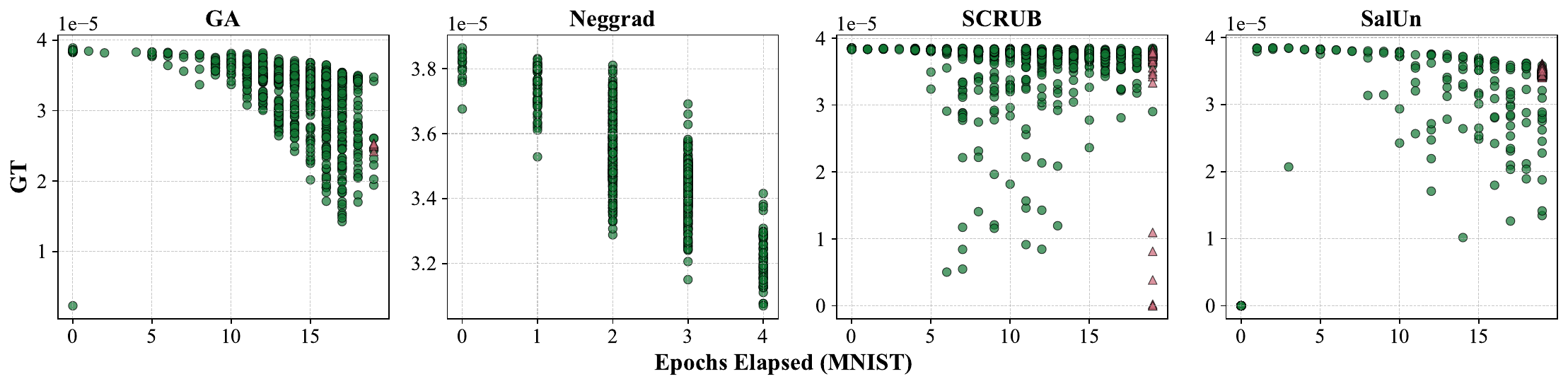}
    \includegraphics[width=\textwidth]{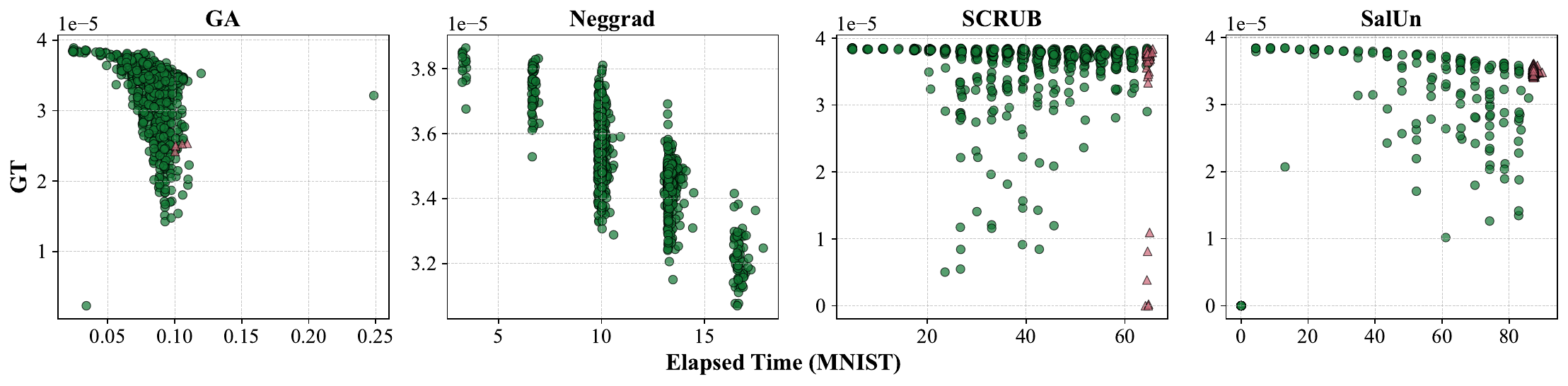}
   \vspace{-5mm}
        \caption{Effectiveness of using Number of Unlearning Epochs (NUE) as index of unlearning difficulty. (Top) Number of epochs used to achieve guaranteed unlearning (flip of decision). (Bottom) Wall clock time elapsed to achieve guaranteed unlearning. The experiments is conducted on ResNet-18 model trained on MNIST dataset.}
    \label{fig:unlearning_steps_mnist}
\end{figure*}

\newpage

\section{Resistance to Membership Inference Attack (MIA)}

\begin{figure*}[t]
    \centering
    \includegraphics[width=\textwidth]{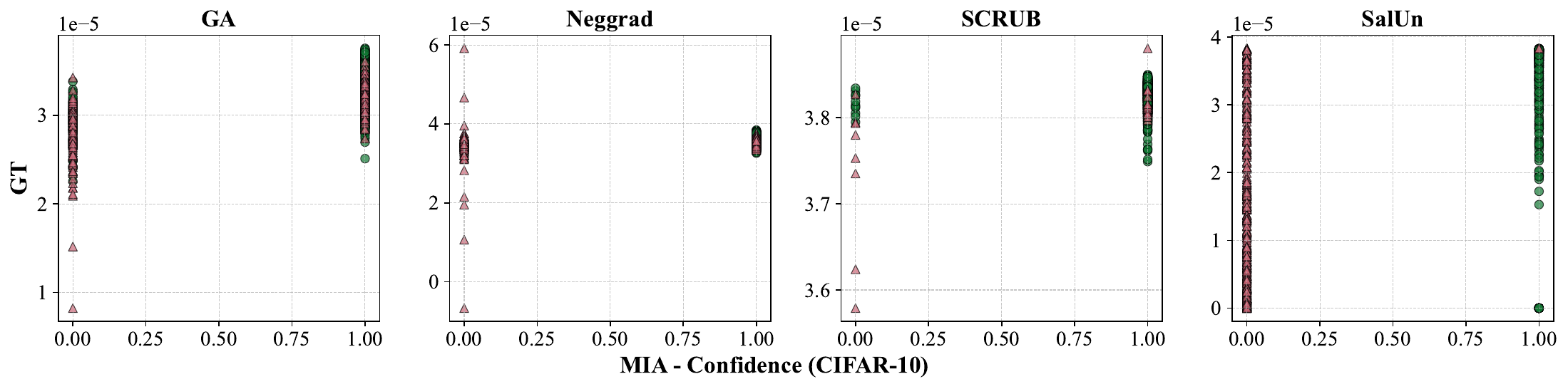}
    \includegraphics[width=\textwidth]{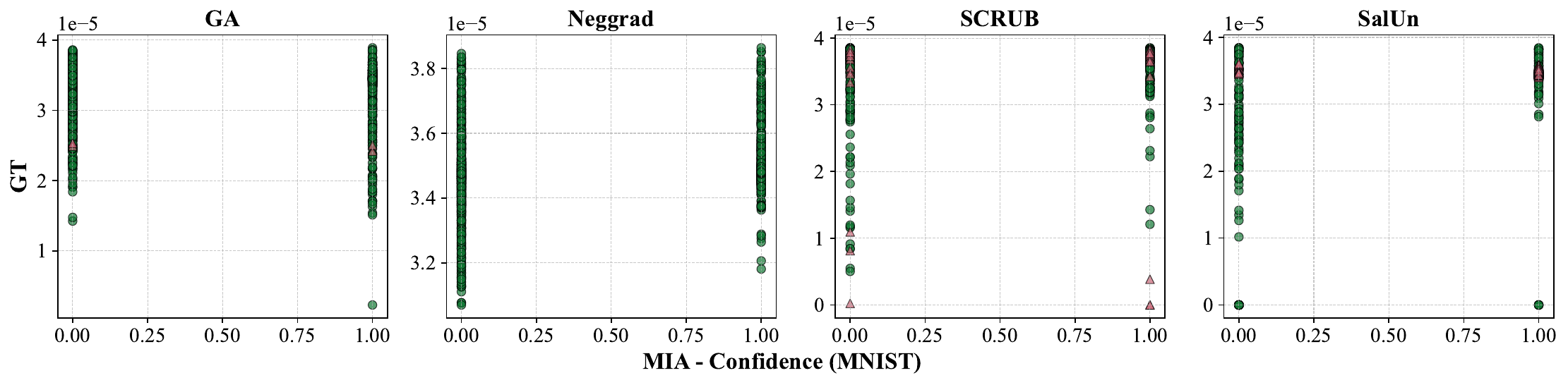}
    \caption{Effectiveness of MIA as index of unlearning difficulty. A prediction of "1" indicates successful unlearning, where the unlearned model no longer retains information about the data, while "0" signifies failed unlearning, where the model still remembers the unlearned samples. The experiments is conducted on ResNet-18 model trained on (from top to bottom) CIFAR10 and MNIST datasets.}
    \label{fig:mia_app}
    \vspace{-0.4cm}
\end{figure*}

\newpage

\section{Dataset Samples}

\begin{figure*}[t]
    \centering
    \includegraphics[width=\textwidth]{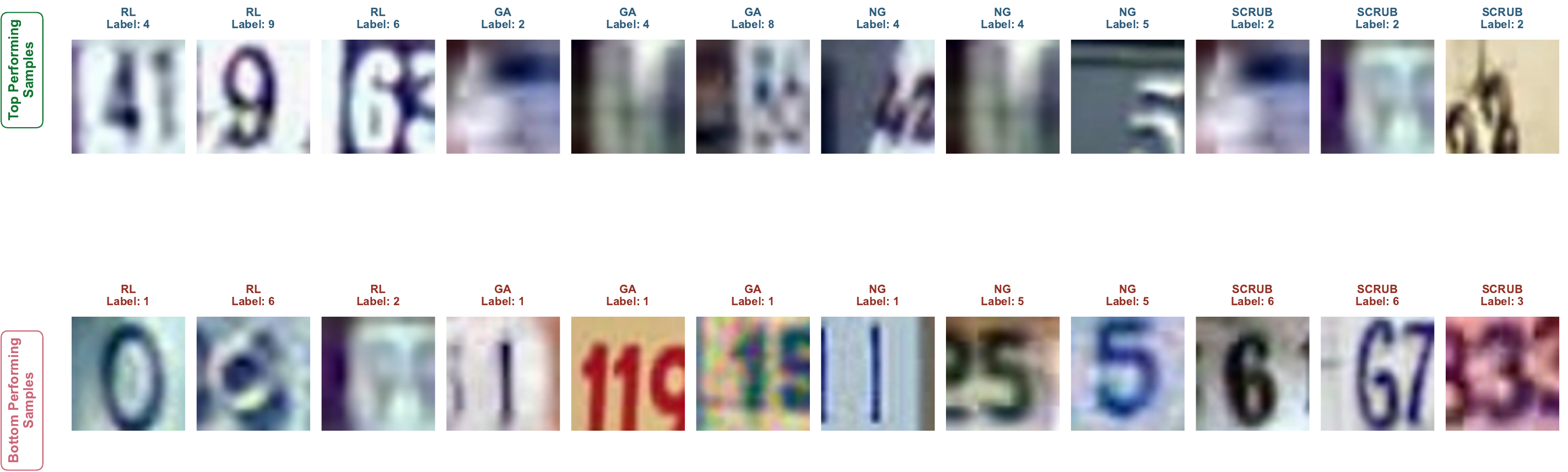}
    \includegraphics[width=\textwidth]{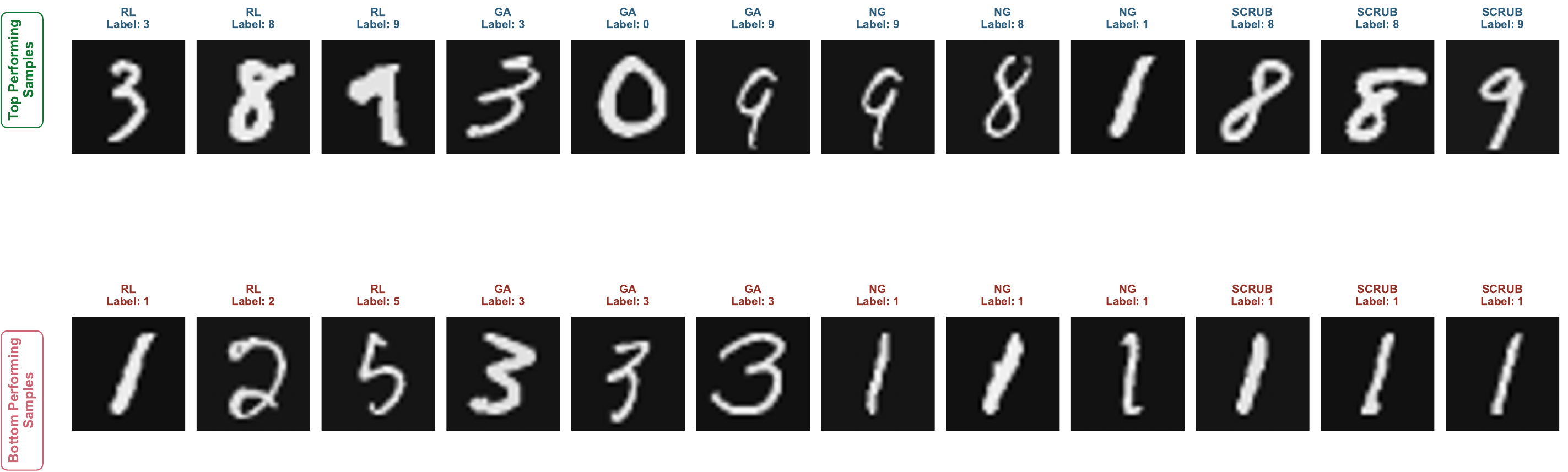}
    \includegraphics[width=\textwidth]{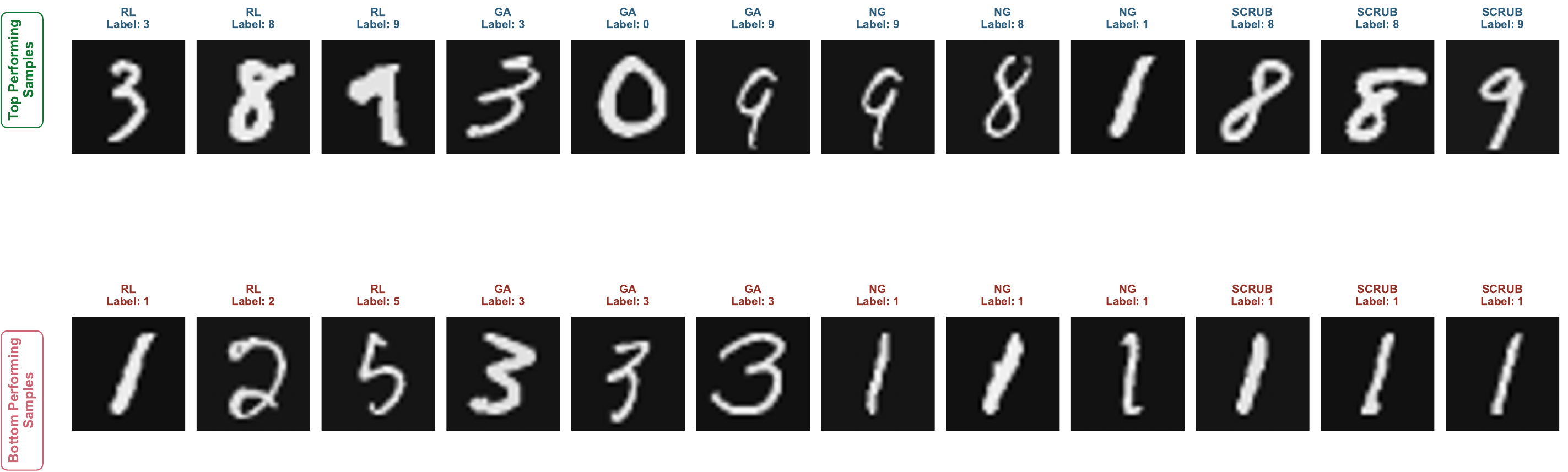}
    \caption{Top-3 easiest and and most difficult samples unlearned by GA, NegGrad, SCRUB and SalUn for machine unlearning flagged by "Test Accuracy" presented in section \ref{sec:six_factors}. In every picture, the top row ("Top Performing Samples") is associated to easy samples and the bottom row ("Bottom Performing Samples") for the difficult samples. The datasets shown (from top to bottom) are SVHN, CIFAR10, and MNIST, respectively.}
    \label{fig:easy_diff_test_acc}
    \vspace{-0.4cm}
\end{figure*}

\begin{figure*}[t]
    \centering
    \includegraphics[width=\textwidth]{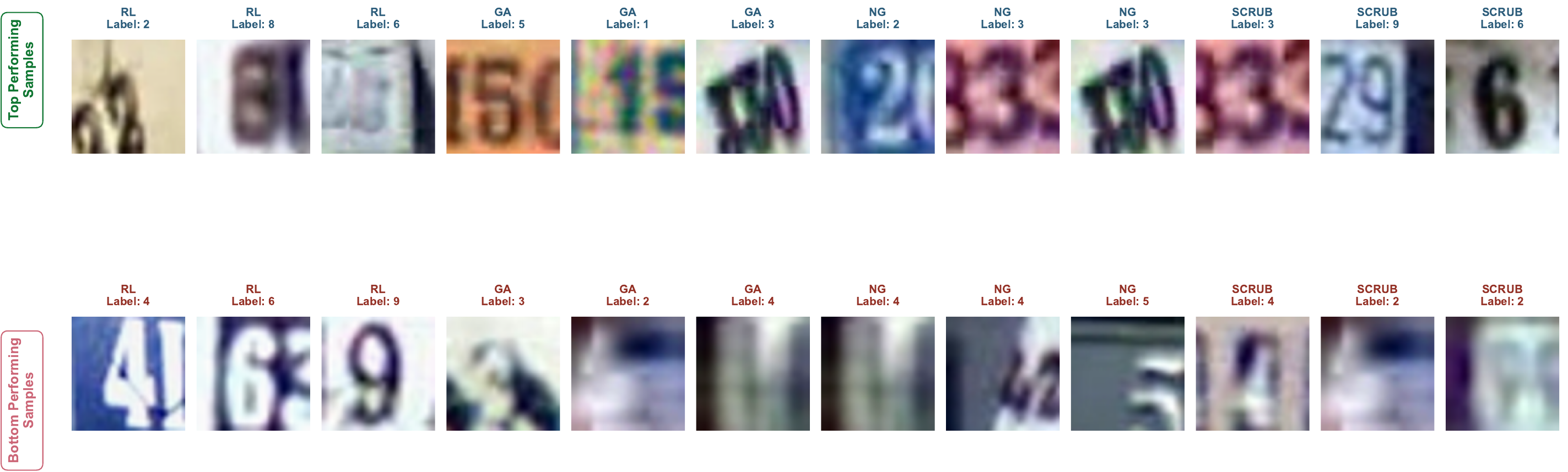}
    \includegraphics[width=\textwidth]{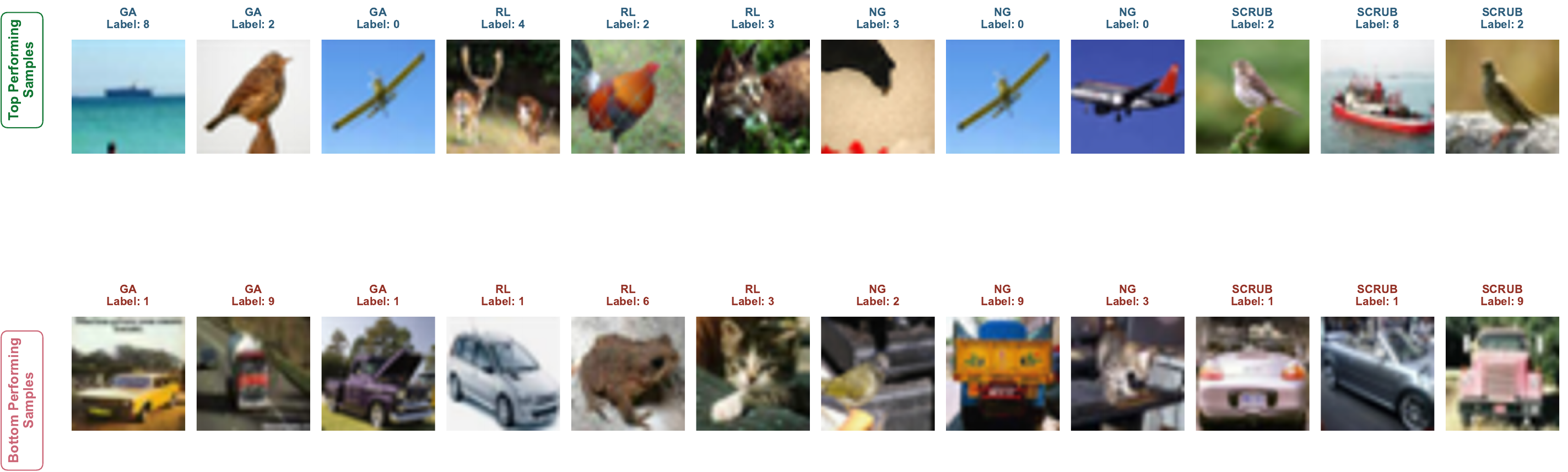}
    \includegraphics[width=\textwidth]{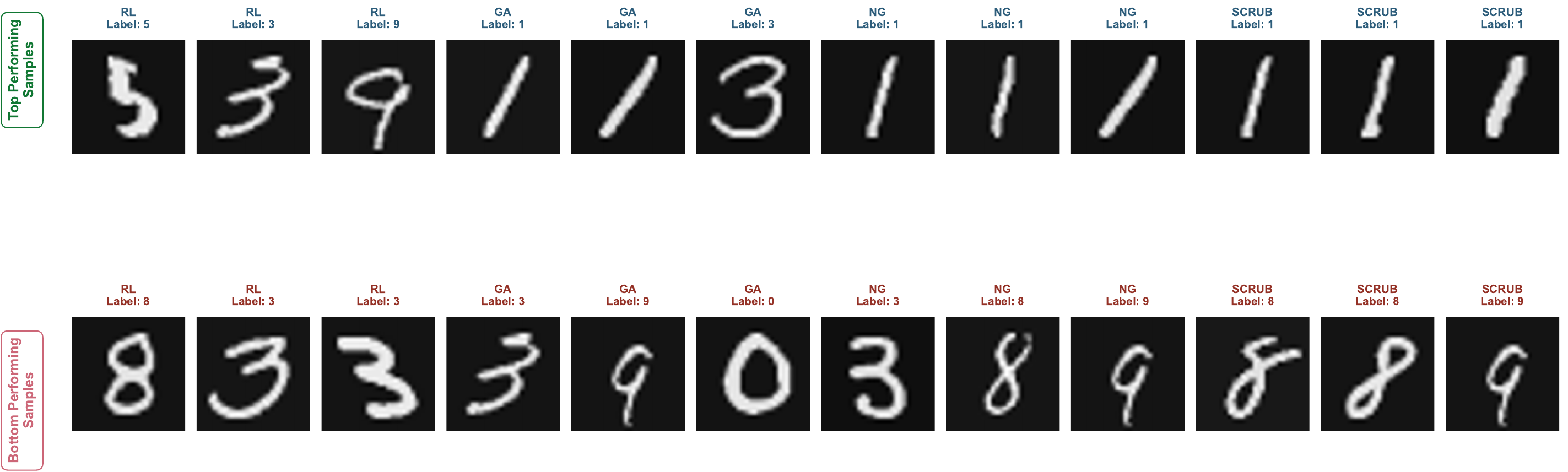}
    \caption{Top-3 easiest and and most difficult samples unlearned by GA, NegGrad, SCRUB and SalUn for machine unlearning flagged by "Test Loss Difference" presented in section \ref{sec:six_factors}. In every picture, the top row ("Top Performing Samples") is associated to easy samples and the bottom row ("Bottom Performing Samples") for the difficult samples. The datasets shown (from top to bottom) are SVHN, CIFAR10, and MNIST, respectively.}
    \label{fig:easy_diff_test_loss}
    \vspace{-0.4cm}
\end{figure*}

\begin{figure*}[t]
    \centering
    \includegraphics[width=\textwidth]{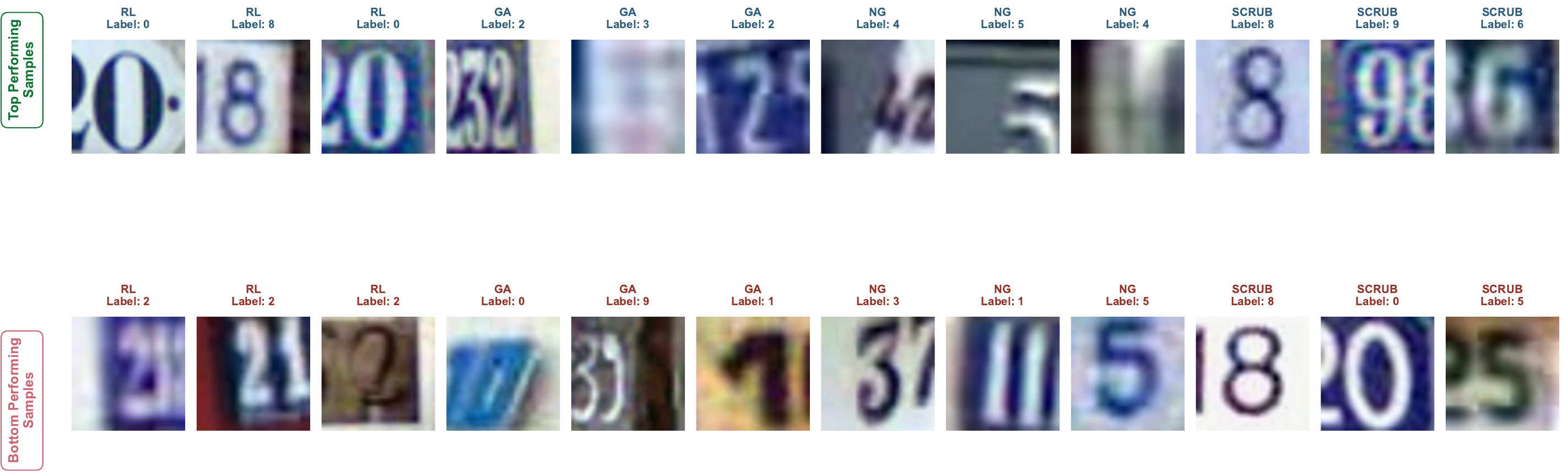}
    \includegraphics[width=\textwidth]{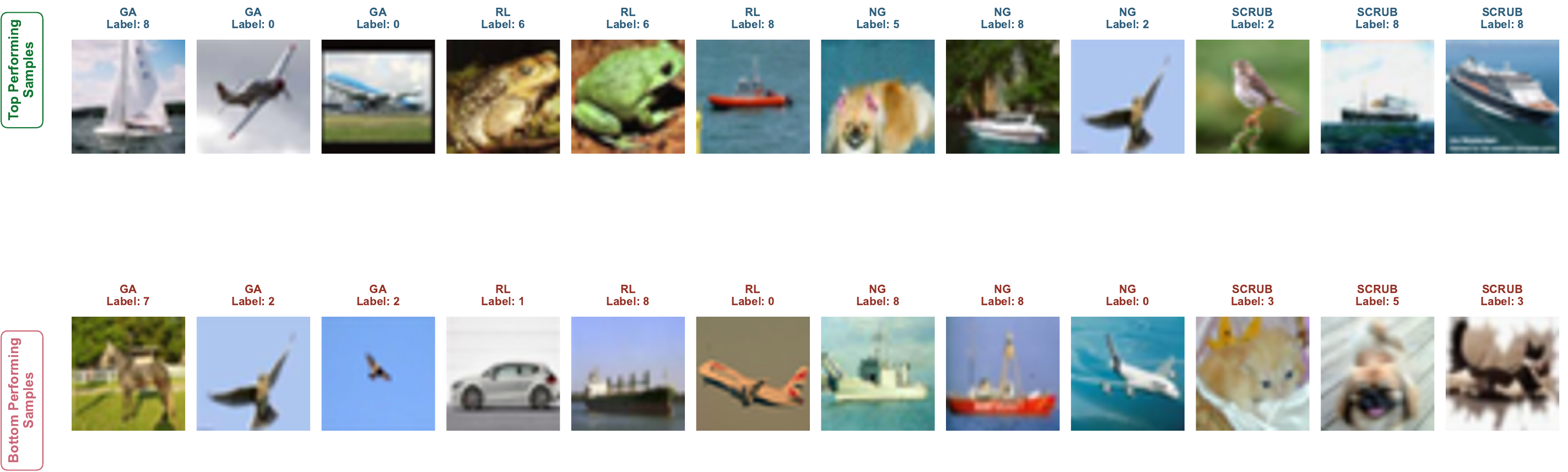}
    \includegraphics[width=\textwidth]{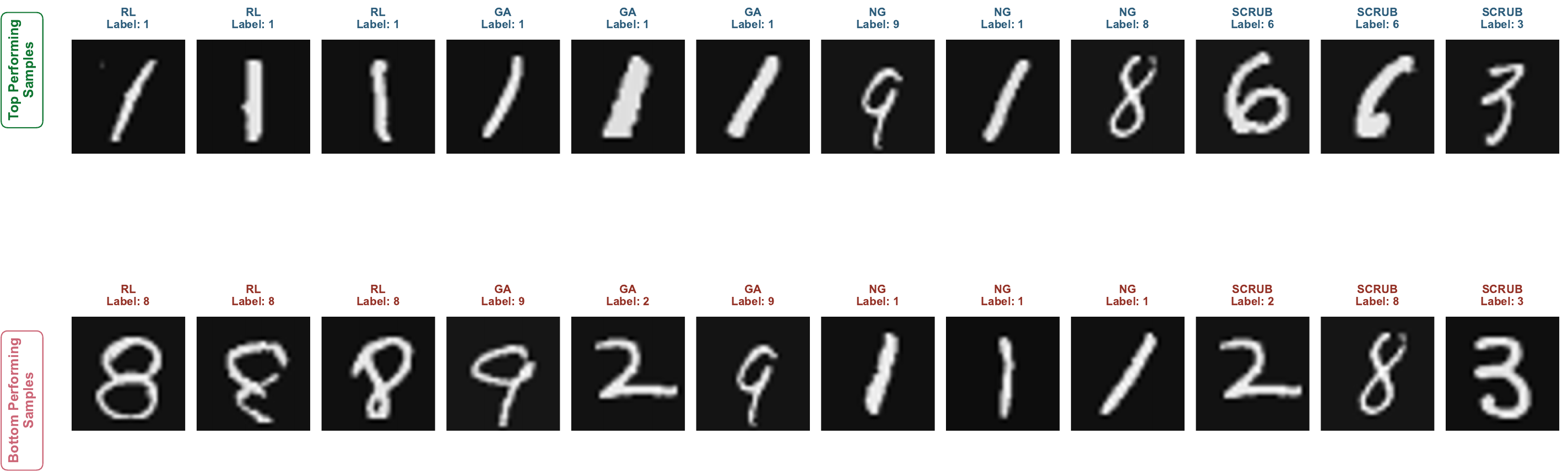}
    \caption{Top-3 easiest and and most difficult samples unlearned by GA, NegGrad, SCRUB and SalUn for machine unlearning flagged by "Forget Loss Difference" presented in section \ref{sec:six_factors}. In every picture, the top row ("Top Performing Samples") is associated to easy samples and the bottom row ("Bottom Performing Samples") for the difficult samples. The datasets shown (from top to bottom) are SVHN, CIFAR10, and MNIST, respectively.}
    \label{fig:easy_diff_forget_loss}
    \vspace{-0.4cm}
\end{figure*}

\begin{figure*}[t]
    \centering
    \includegraphics[width=\textwidth]{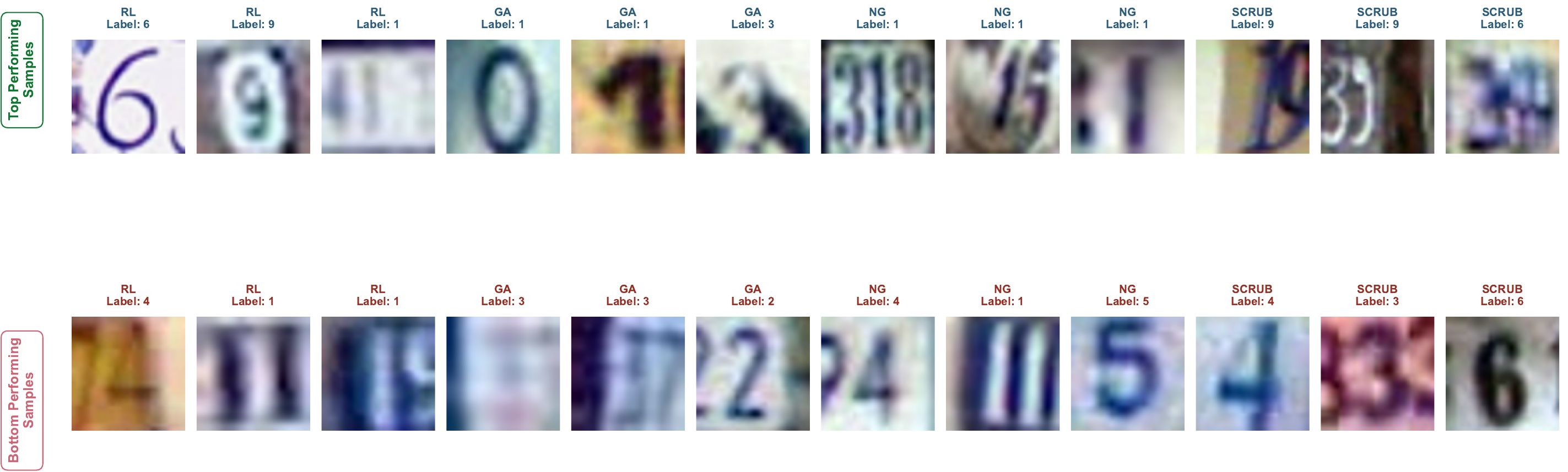}
    \includegraphics[width=\textwidth]{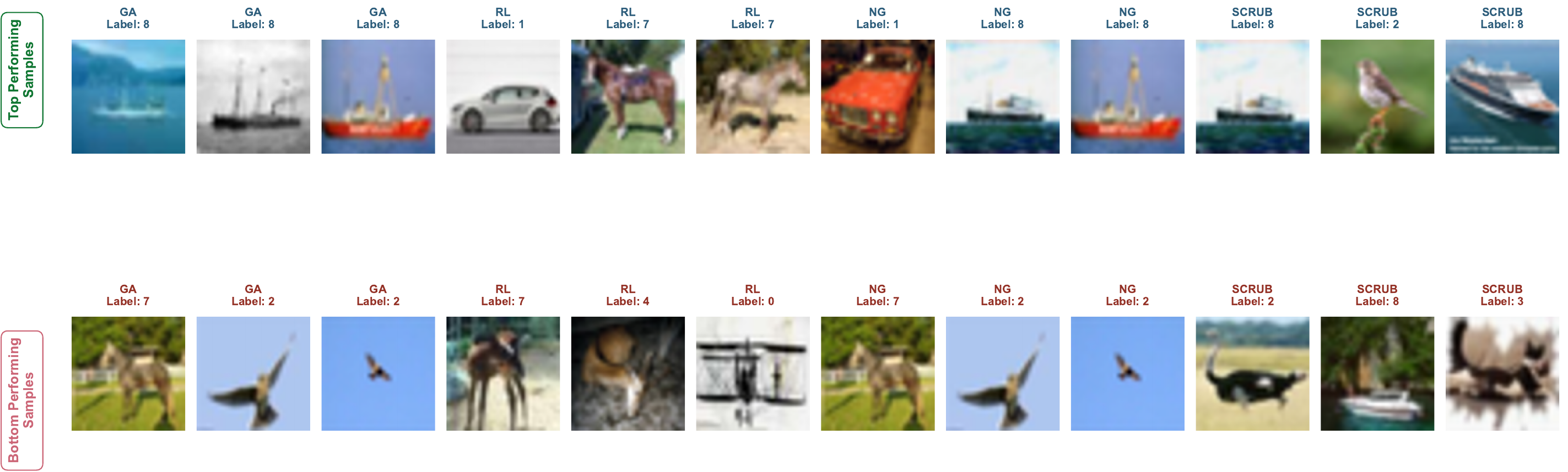}
    \includegraphics[width=\textwidth]{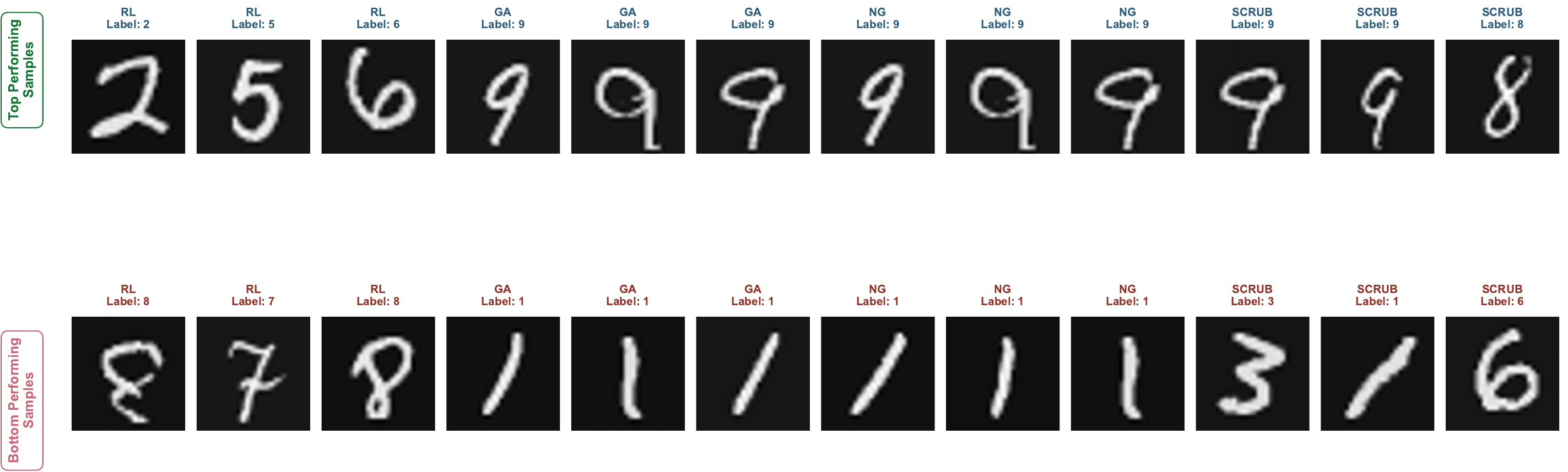}
    \caption{Top-3 easiest and and most difficult samples unlearned by GA, NegGrad, SCRUB and SalUn for machine unlearning flagged by "Model Distance - Absolute" presented in section \ref{sec:six_factors}. In every picture, the top row ("Top Performing Samples") is associated to easy samples and the bottom row ("Bottom Performing Samples") for the difficult samples. The datasets shown (from top to bottom) are SVHN, CIFAR10, and MNIST, respectively.}
    \label{fig:easy_diff_dist_abs}
    \vspace{-0.4cm}
\end{figure*}

\begin{figure*}[t]
    \centering
    \includegraphics[width=\textwidth]{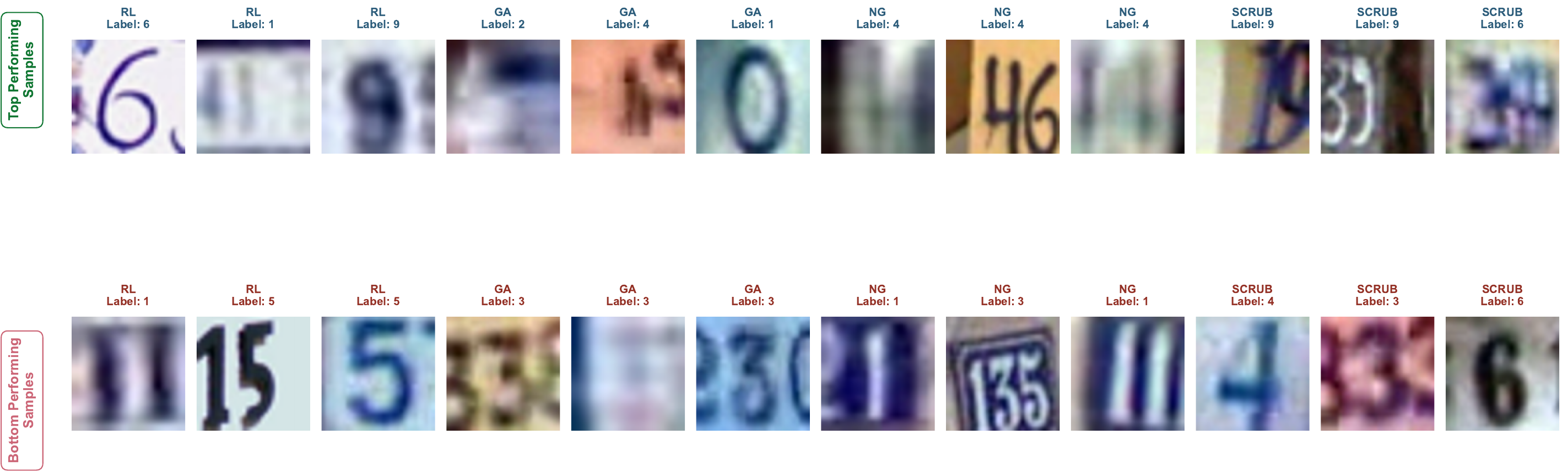}
    \includegraphics[width=\textwidth]{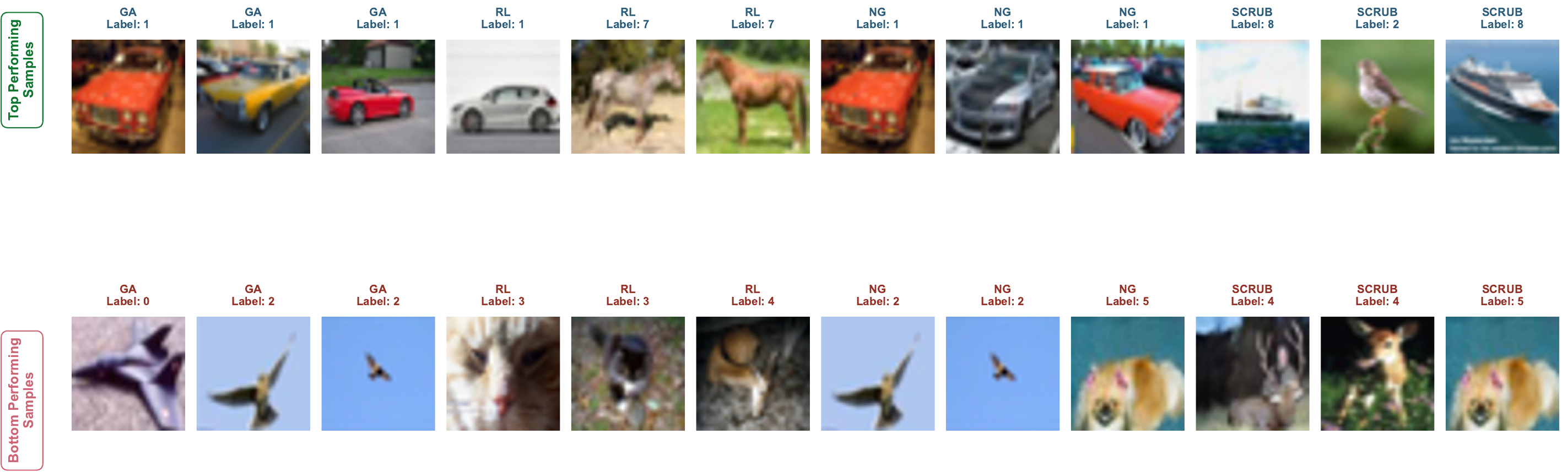}
    \includegraphics[width=\textwidth]{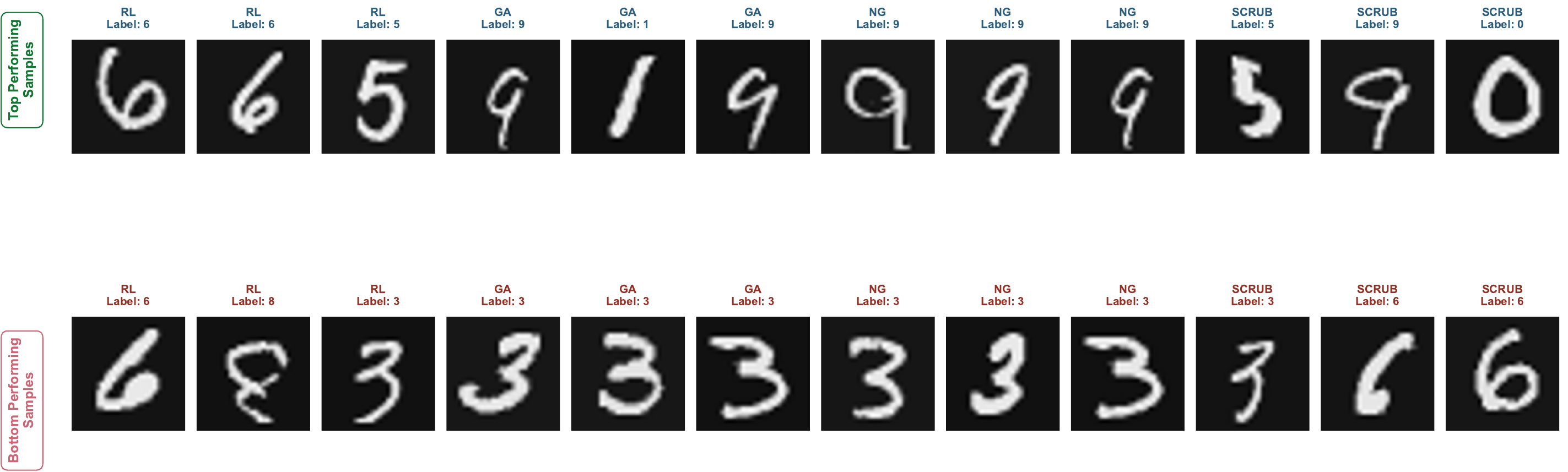}
    \caption{Top-3 easiest and and most difficult samples unlearned by GA, NegGrad, SCRUB and SalUn for machine unlearning flagged by "Model Distance - KL" presented in section \ref{sec:six_factors}. In every picture, the top row ("Top Performing Samples") is associated to easy samples and the bottom row ("Bottom Performing Samples") for the difficult samples.  The datasets shown (from top to bottom) are SVHN, CIFAR10, and MNIST, respectively.}
    \label{fig:easy_diff_dist_kl}
    \vspace{-0.4cm}
\end{figure*}

\begin{figure*}[t]
    \centering
    \includegraphics[width=\textwidth]{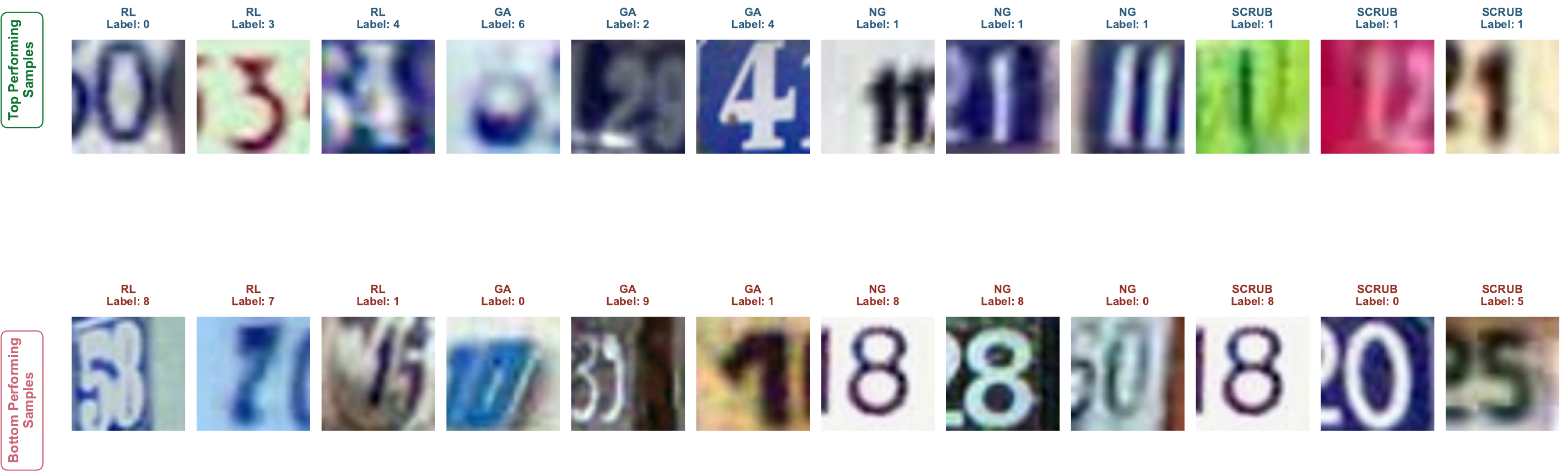}
    \includegraphics[width=\textwidth]{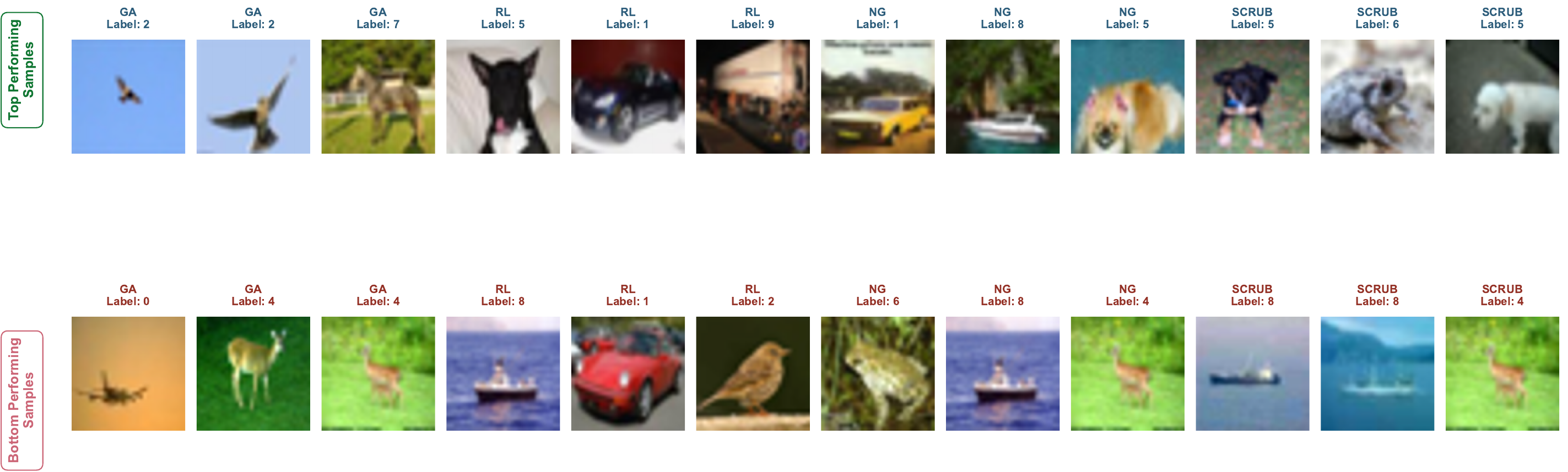}
    \includegraphics[width=\textwidth]{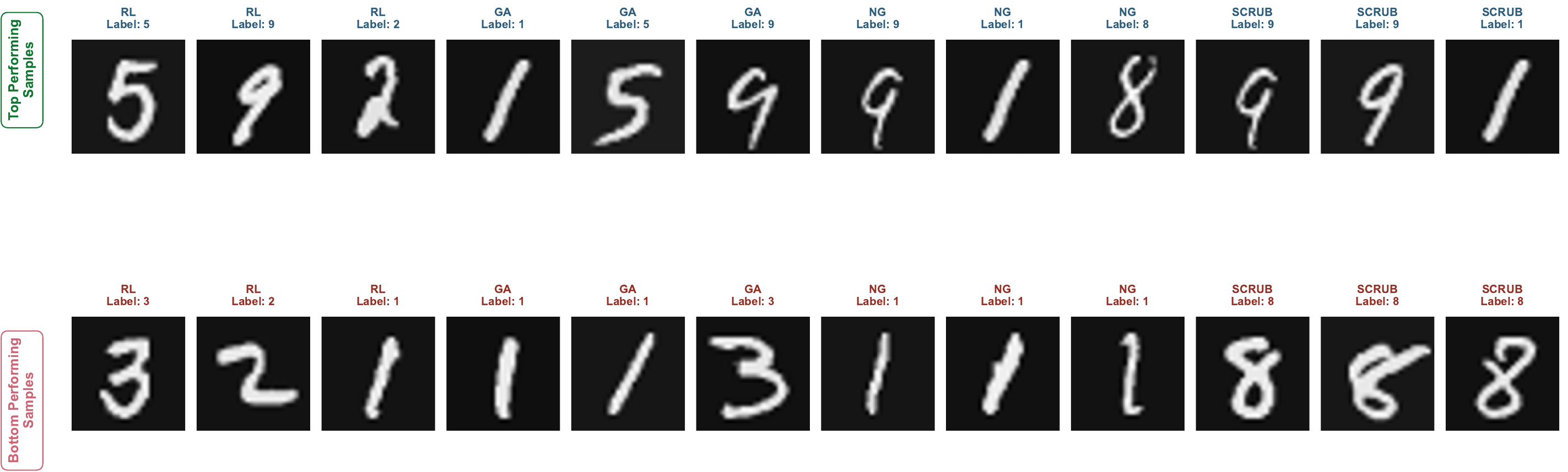}
    \caption{Top-3 easiest and and most difficult samples unlearned by GA, NegGrad, SCRUB and SalUn for machine unlearning flagged by "Epoch Elapsed" presented in section \ref{sec:six_factors}. In every picture, the top row ("Top Performing Samples") is associated to easy samples and the bottom row ("Bottom Performing Samples") for the difficult samples. The datasets shown (from top to bottom) are SVHN, CIFAR10, and MNIST, respectively.}
    \label{fig:easy_diff_epoch}
    \vspace{-0.4cm}
\end{figure*}

\begin{figure*}[t]
    \centering
    \includegraphics[width=\textwidth]{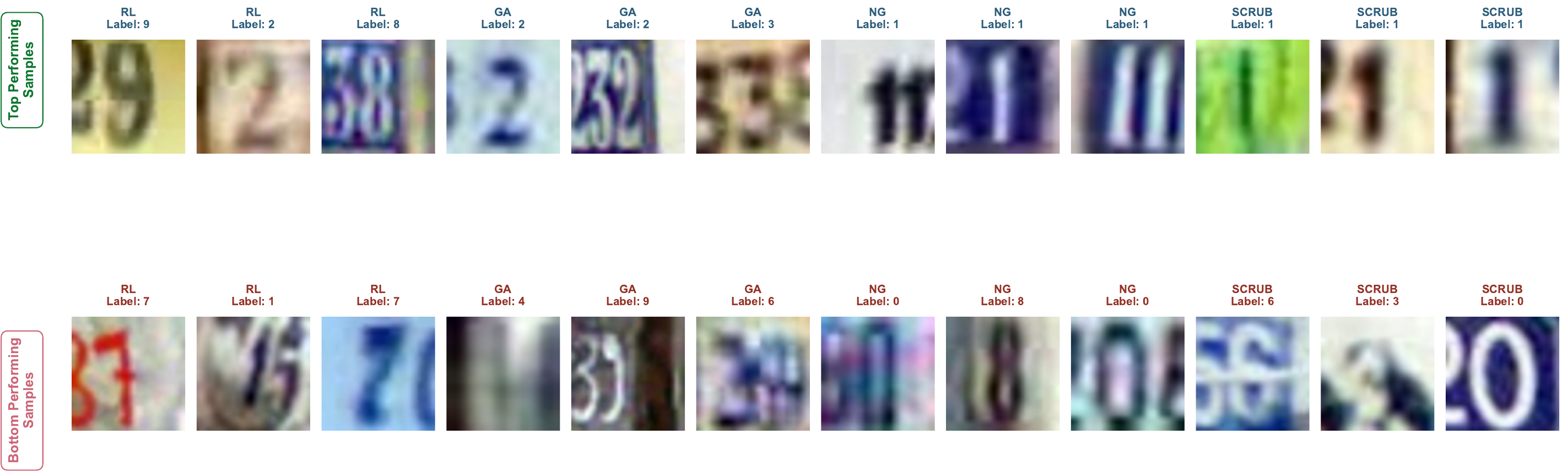}
    \includegraphics[width=\textwidth]{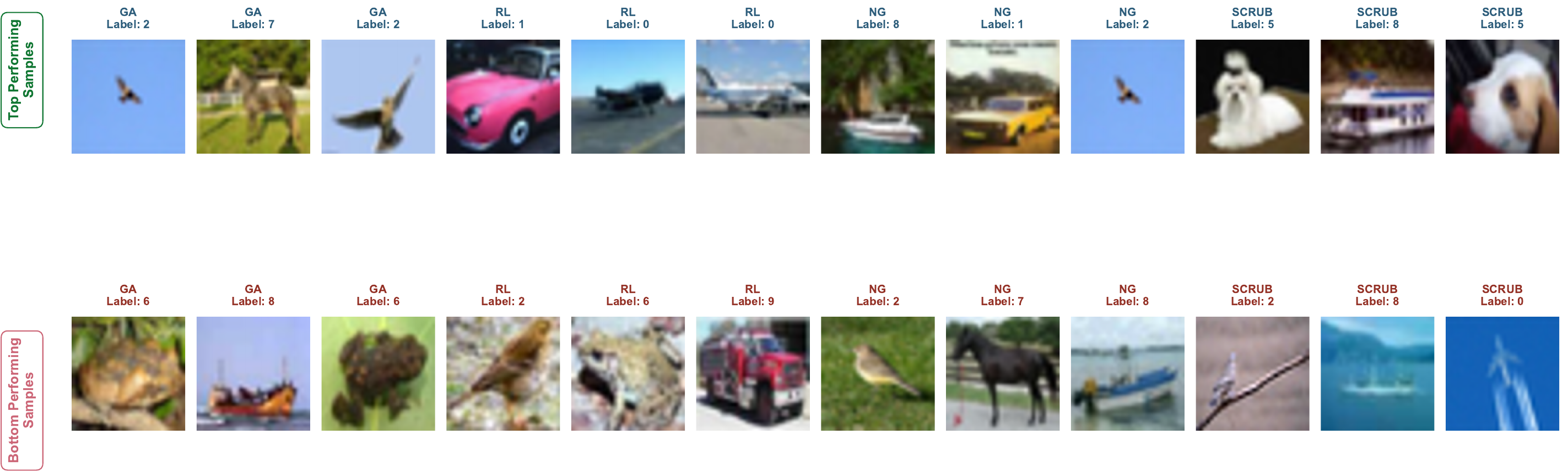}
    \includegraphics[width=\textwidth]{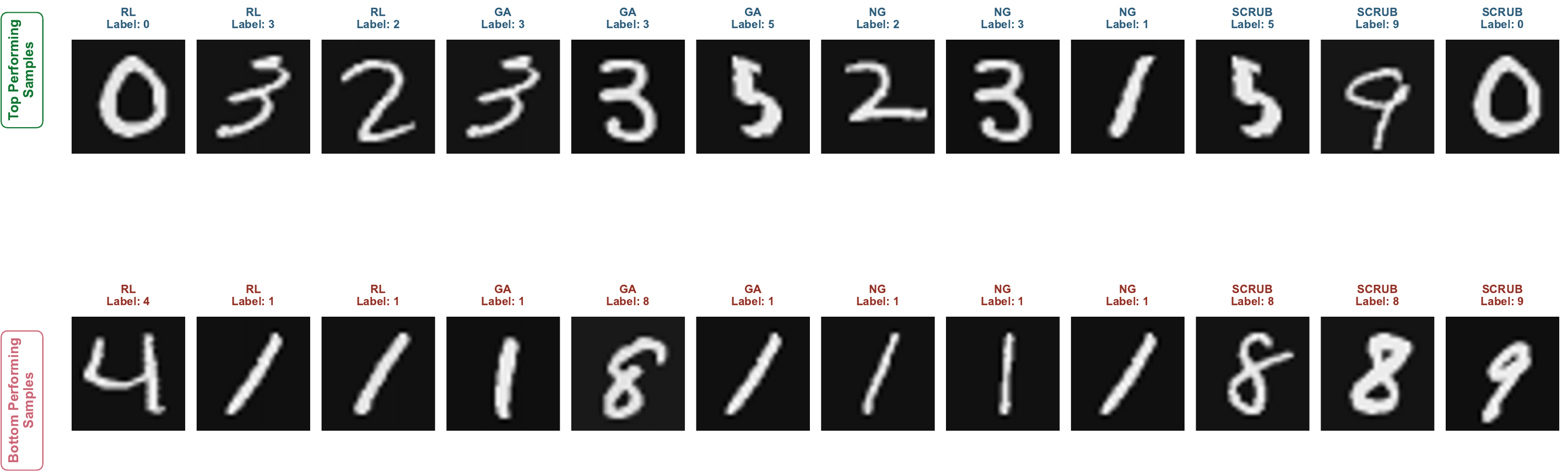}
    \caption{Top-3 easiest and and most difficult samples unlearned by GA, NegGrad, SCRUB and SalUn for machine unlearning flagged by "Time Elapsed" presented in section \ref{sec:six_factors}. In every picture, the top row ("Top Performing Samples") is associated to easy samples and the bottom row ("Bottom Performing Samples") for the difficult samples. The presence of similar samples suggests that this factor identifies the easiest and most difficult cases independently of the unlearning algorithm. The datasets shown (from top to bottom) are SVHN, CIFAR10, and MNIST, respectively.}
    \label{fig:easy_diff_time}
    \vspace{-0.4cm}
\end{figure*}

\begin{figure*}[t]
    \centering
    \includegraphics[width=\textwidth]{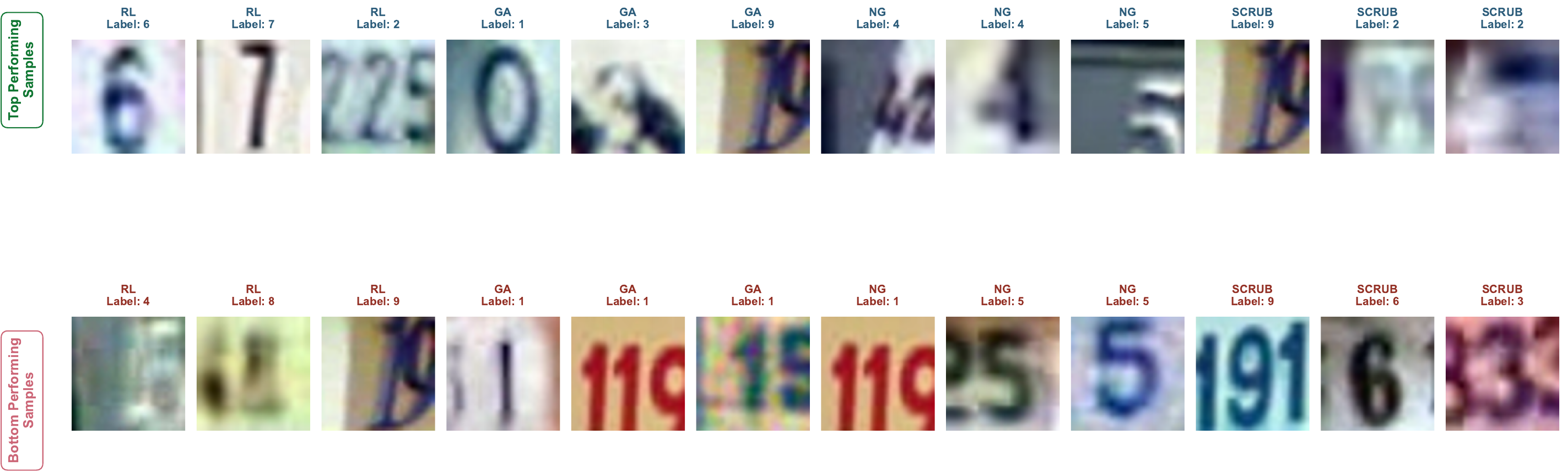}
    \includegraphics[width=\textwidth]{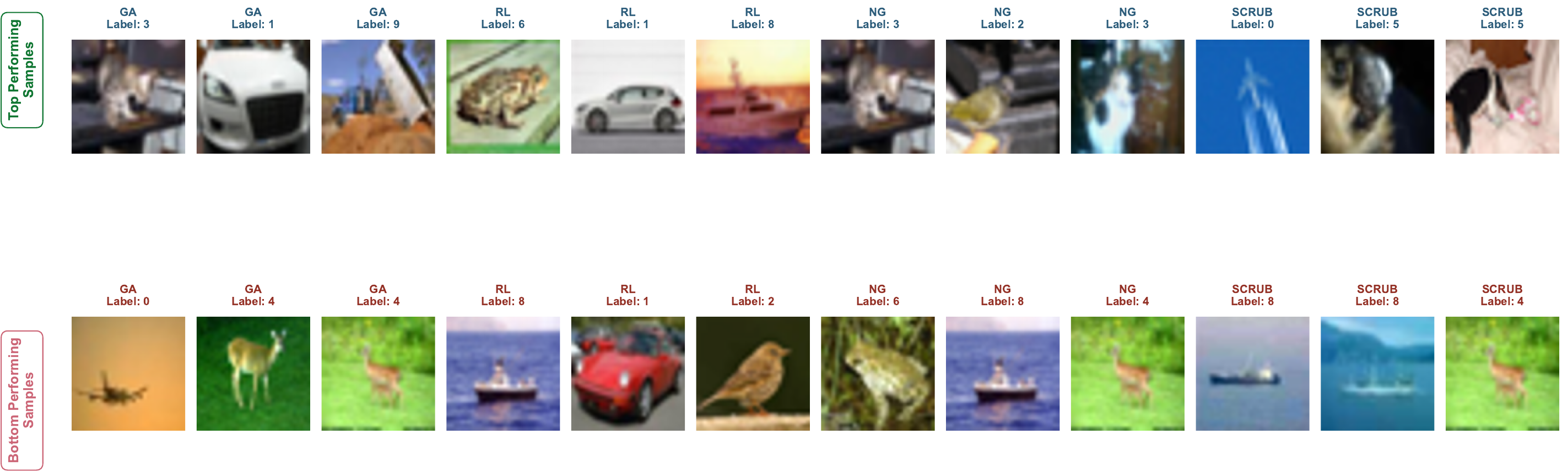}
    \includegraphics[width=\textwidth]{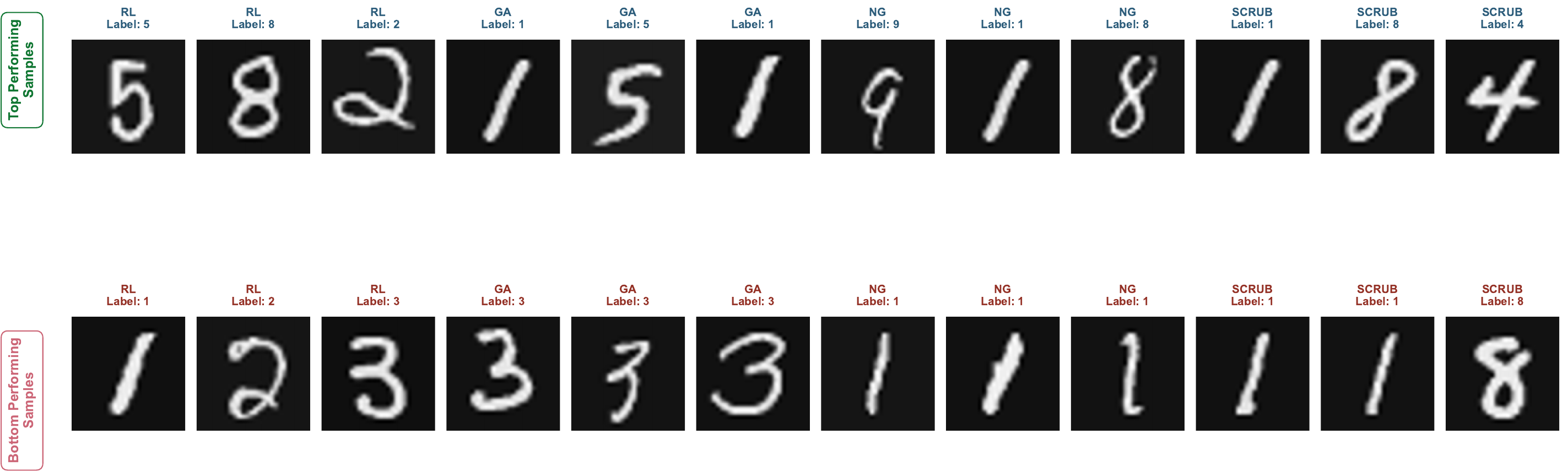}
    \caption{Top-3 easiest and and most difficult samples unlearned by GA, NegGrad, SCRUB and SalUn for machine unlearning flagged by "MIA" presented in section \ref{sec:six_factors}. In every picture, the top row ("Top Performing Samples") is associated to easy samples and the bottom row ("Bottom Performing Samples") for the difficult samples. The datasets shown (from top to bottom) are SVHN, CIFAR10, and MNIST, respectively.}
    \label{fig:easy_diff_mia}
    \vspace{-0.4cm}
\end{figure*}

\end{document}